%% file: main.tex
\documentclass[10pt,twocolumn,letterpaper]{article}

\usepackage{cvpr}              

\definecolor{cvprblue}{rgb}{0.21,0.49,0.74}
\usepackage[pagebackref,breaklinks,colorlinks,allcolors=cvprblue]{hyperref}

\def\paperID{*****} 
\def\confName{CVPR}
\def\confYear{2025}

\title{Bringing CLIP to the Clinic: Dynamic Soft Labels and Negation-Aware Learning for Medical Analysis}

\author{Hanbin Ko\textsuperscript{1,2}, Chang-Min Park\textsuperscript{1,2,3}\thanks{Corresponding author}\
\\[0.2em]
\textsuperscript{1}Interdisciplinary Program in Bioengineering, Seoul National University Graduate School,\\
\textsuperscript{2}Integrated Major in Innovative Medical Science, Seoul National University Graduate School,\\
\textsuperscript{3}Department of Radiology, Seoul National University Hospital\\[0.2em]
{\tt\small \{lucasko1994, morphius\}@snu.ac.kr}
}

\begin{document}
\maketitle

\input{sec/0_abstract}    
\input{sec/1_intro}

\input{sec/2_related_works}
\input{sec/3_cxralign}
\input{sec/4_method}

\input{sec/5_experiment}

\input{sec/6_discussion}
\input{sec/7_conclusion}
{
    \small
    \bibliographystyle{ieeenat_fullname}
    \bibliography{main}
}

\input{sec/X_suppl}

\end{document}

%% file: sec/0_abstract.tex
\begin{abstract}
The development of large-scale image-text pair datasets has significantly advanced self-supervised learning in Vision-Language Processing (VLP). However, directly applying general-domain architectures such as CLIP to medical data presents challenges, particularly in handling negations and addressing the inherent data imbalance of medical datasets. To address these issues, we propose a novel approach that integrates clinically-enhanced dynamic soft labels and medical graphical alignment, thereby improving clinical comprehension and improving the applicability of contrastive loss in medical contexts. Furthermore, we introduce negation-based hard negatives to deepen the model’s understanding of the complexities of clinical language. Our approach is easily integrated into medical CLIP training pipeline and achieves state-of-the-art performance across multiple tasks, including zero-shot, fine-tuned classification and report retrieval. To comprehensively evaluate our model's capacity in understanding clinical language, we introduce \textbf{CXR-Align}, a benchmark uniquely designed to evaluate the understanding of negation and clinical information within chest X-ray (CXR) datasets. Experimental results demonstrate that our proposed methods are straightforward to implement and generalize effectively across contrastive learning frameworks, enhancing medical VLP capabilities and advancing clinical language understanding in medical imaging.
\end{abstract}

%% file: sec/1_intro.tex
\section{Introduction}
\label{sec:intro}
CLIP~\cite{clip} has revolutionized Vision-Language Processing (VLP), with particularly promising applications in medical imaging analysis~\cite{clipsurvey}. Medical imaging, especially in areas requiring specialized annotation expertise, greatly benefits from CLIP’s ability to leverage image-text pairs without extensive labeled data, thus enabling efficient representation learning. Consequently, research has increasingly focused on adapting CLIP-like models for CXR data, which is rich in image-report pairs and well-suited to contrastive learning.

\begin{figure}[t]
    \centering
    \includegraphics[width=1\linewidth]{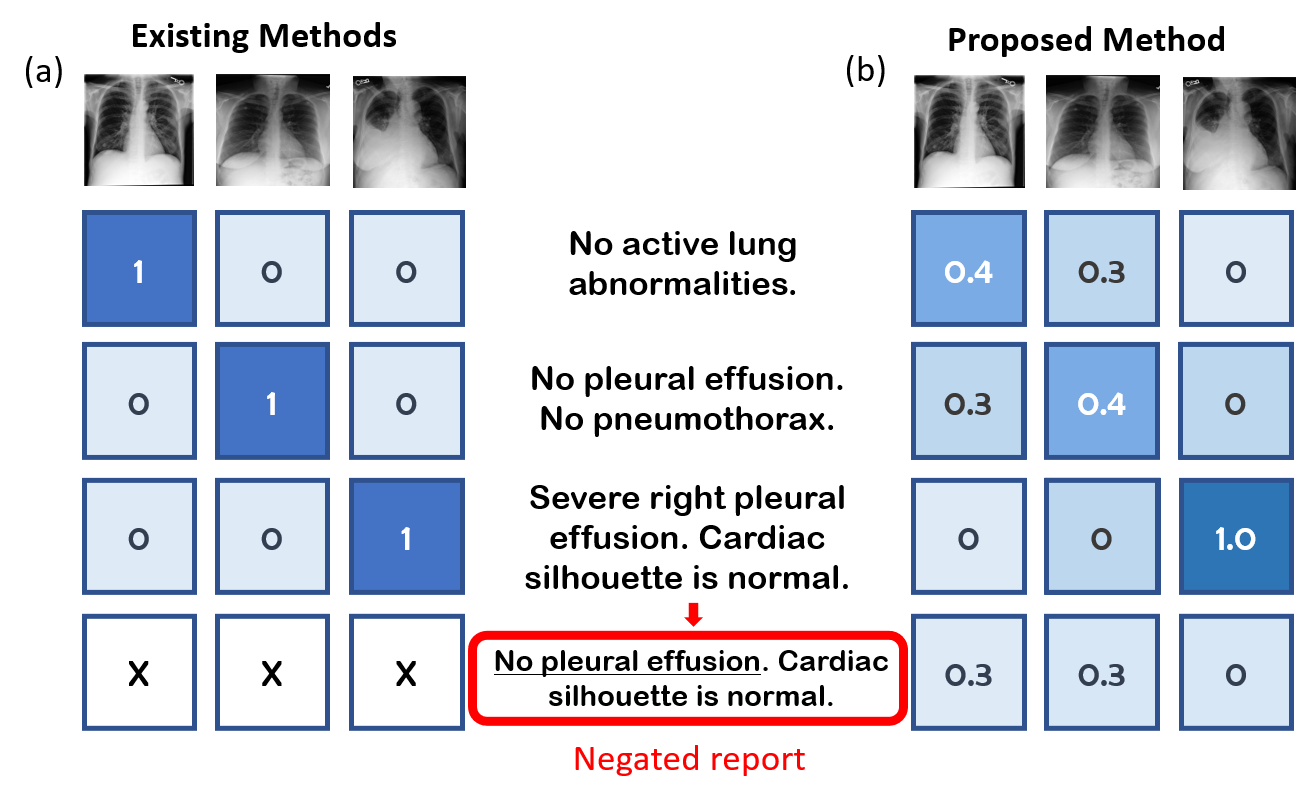}
    \caption{(a) Standard visual-language pre-training approaches using contrastive learning (e.g., InfoNCE). (b) Our approach, leveraging unique medical domain characteristics (e.g., imbalance and negations), dynamically generates soft labels based on clinical, textual, and relational similarities while integrating negations as hard negatives.}
    \label{fig:motivation}
\end{figure}

However, directly applying CLIP’s contrastive learning framework to medical data poses significant challenges due to unique characteristics of medical image-text data. For example, medical reports often contain negations and are subject to considerable data imbalance. While several adaptations, such as Xlip~\cite{wu2024xlip}, CXR-CLIP~\cite{cxrclip}, GLORIA~\cite{huang2021gloria}, BioViL~\cite{biovil}, MedKlip~\cite{wu2023medklip}, MLIP~\cite{mlip}, have sought to improve image-report alignment in the medical field, many overlook specific aspects of CXR reports, such as references to interval changes requiring temporal context from prior images. BioViL-T~\cite{biovilt} addresses this challenge by incorporating prior images to account for temporal considerations.

\begin{table}[h]
    \centering
    \begin{tabular}{l c}
        \toprule
        Report & Count \\
        \midrule
        No acute cardiopulmonary process & 15829 \\
        No acute intrathoracic process & 4643 \\
        No acute cardiopulmonary abnormality & 4488 \\
        No evidence of acute cardiopulmonary process & 1255 \\
        No evidence of pneumonia & 1014 \\
        \bottomrule
    \end{tabular}
    \caption{Top most frequent reports and their counts in the MIMIC training data. Due to the frequent use of predefined templates, duplicated reports amplify imbalance.}
    \label{tab:report_counts}
\end{table}

In addition, radiologists routinely document both the presence and absence of findings (e.g., “No pneumothorax”), making negation a critical feature for precise clinical communication. However, CLIP-based models, which often exhibit "bag-of-words" characteristics~\cite{bagofwords}, can struggle to fully interpret negated terms~\cite{negation}. For effective comprehension of CXR reports, it is essential for the model to correctly understand the purpose and implications of these negated entities.

Contrastive learning in medical datasets also contends with significant imbalance. In general domains, increasing batch sizes during CLIP training improves gradient estimation by introducing a wider variety of negative samples~\cite{simclr, clip}. However, medical datasets are heavily skewed towards normal cases and exhibit template-based duplication, as illustrated in ~\cref{tab:report_counts}. Radiologists frequently use predefined templates for routine findings, resulting in numerous near-duplicate reports that amplify data imbalance. In this context, larger batch sizes increase the likelihood of semantically identical or duplicate reports being treated as negatives, introducing noise that conflicts with the objectives of contrastive learning. Although the ICU-focused MIMIC~\cite{mimic3, mimic4} dataset provides some diversity, such imbalances can be even more pronounced in general hospital datasets where templated language and normal cases are prevalent, comprising over three-quarters of the data.

In this study, we address the challenges of data imbalance and negation specifically within the context of CLIP for medical datasets. Unlike traditional medical imbalance issues, this imbalance arises uniquely in contrastive settings, where solutions like report rewriting \cite{rewrite} are insufficient. To our knowledge, this is the first paper to directly tackle these medical-specific features common in clinical reports at a global-feature scale. We define data imbalance as primarily semantic overlap within the batch, often but not exclusively limited to normal CXR reports.

Our approach focuses on single-image scenarios, leaving temporal considerations for future work. To mitigate imbalance, we introduce a \textit{clinically-enhanced dynamic soft-labeling} method that incorporates clinical and textual similarity into contrastive learning, allowing the model to better interpret the clinical relationship of reports in each batch. For handling negations, we generate negation-based hard negatives to enhance CLIP training, strengthening the model’s capacity to create accurate clinical representations. Additionally, we integrate \textit{graph embeddings} to enrich the image-text architecture, capturing domain-specific relationship that result in stable learning and improved performance across tasks such as zero-shot and fine-tuned classification, adversarial prediction, CXR-report alignment through negations and clinical entities, normal case detection, and report retrieval.

\begin{itemize}[leftmargin=*] \item We propose a contrastive learning method that leverages dynamic soft labels, incorporating clinical and textual similarity, to address data imbalance and improve training stability in medical settings. \item We introduce the \textit{CXR-Align} benchmark, designed to evaluate models on negation handling and clinical alignment, advancing the assessment of medical VLP models. \item We create a negation generation pipeline to synthesize hard negatives, strengthening the model’s understanding of negated findings, which works synergistically with dynamic soft labels. \item We integrate graph embeddings into the contrastive framework to capture the unique characteristics of medical data, refining soft-labels and enhancing negation comprehension. \item Our method demonstrates strong performance across tasks like classification, adversarial prediction, CXR-report alignment, normal case detection and retrieval, surpassing baseline and state-of-the-art CLIP-based models. \end{itemize}

%% file: sec/2_related_works.tex
\begin{figure*}
  \centering
  \includegraphics[width=1.0\linewidth]{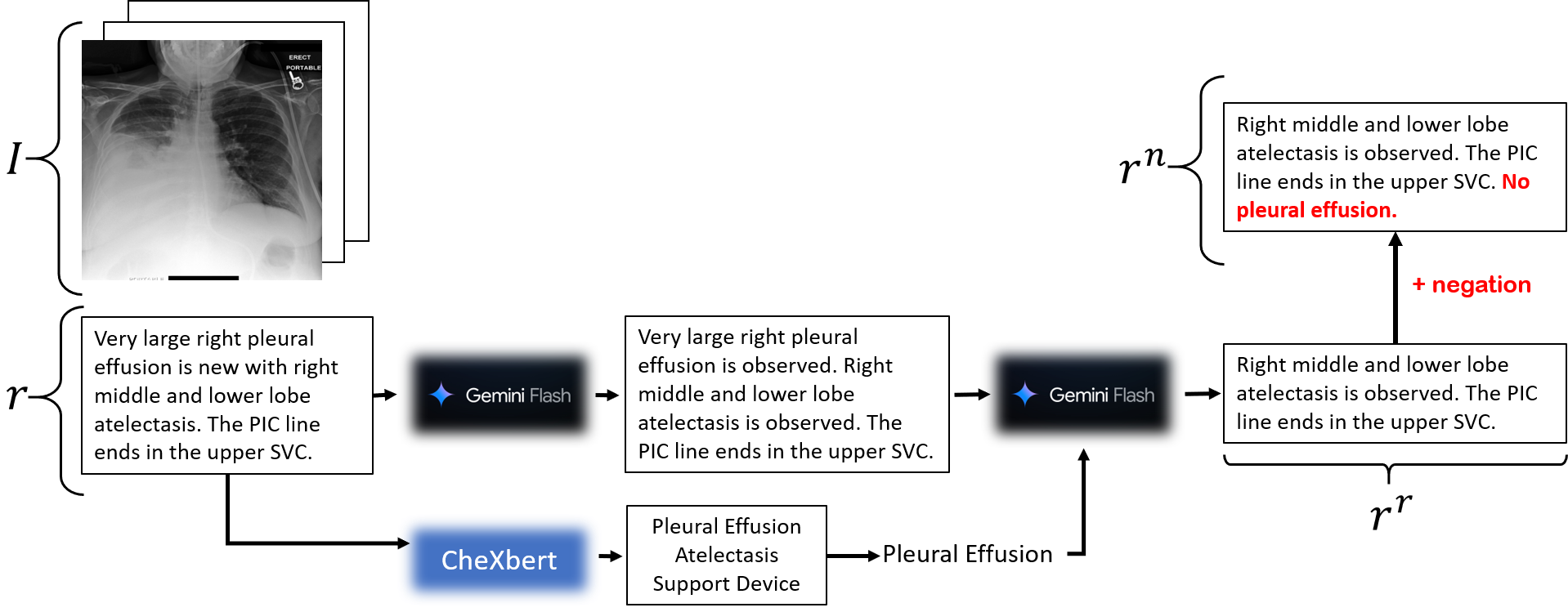}
  \caption{Given a CXR report, CheXbert identifies all positive entities, and one is randomly selected. A language model then (i) splits the report so each sentence contains a single clinical entity without temporal statements and (ii) removes sentences related to the selected entity. Finally, a negation for the selected entity is added at a random position within the report (beginning, middle, or end).}
  \label{fig:cxrneg}
\end{figure*}

\section{Related Works}

\subsection{Medical VLP for Chest X-Rays} 
Recently, contrastive learning approaches inspired by CLIP~\cite{clip} have gained traction in medical applications, benefiting from the abundant paired data in CXR tasks~\cite{chexpertplus, mimic3}. Notable models include CheXzero~\cite{chexzero} and ConViRT~\cite{convirt}, which align image and text representations trained on the MIMIC dataset, and GLORIA~\cite{huang2021gloria}, which employs local representations for fine-grained alignment. CXR-CLIP~\cite{cxrclip} explores image-to-image alignment, while XLiP~\cite{wu2024xlip} and BioViL~\cite{biovil} adopt masked modeling to predict masked elements in both images and text. BioViL-T~\cite{biovilt} uniquely incorporates temporal information using prior images to capture interval changes in CXR reports. MedKlip~\cite{wu2023medklip} and MLIP~\cite{mlip} incorporate clinical knowledge to enhance the models with domain-specific information. Notably, MLIP highlighted that semantic overlap within batches can cause problems in contrastive learning settings, proposing a solution that uses external knowledge to bind similar semantics in a local-scale environment. Despite these advancements, challenges such as data imbalance and frequent negations in medical text still remain largely unresolved, especially on a global scale, hindering the development of reliable clinical models.

\subsection{CLIP for Compositional Understanding and Negations} 
Assessing the compositional capabilities of vision-language models like CLIP is essential for evaluating their generalization to new combinations of visual and textual information. The CREPE benchmark~\cite{crepe} introduced metrics revealing that large-scale pretraining often falls short in compositional reasoning. Yuksekgonul \etal~\cite{bagofwords} highlighted that CLIP often behaves like a "bag-of-words" model, raising concerns about its textual comprehension. SUGARCREPE~\cite{sugarcrepe} addresses these biases by generating fluent and plausible hard negatives through language models with adversarial refinement. To enhance negation handling, CoN-CLIP~\cite{negation} achieves strong results on the CC-Neg benchmark, underscoring the importance of handling negations in VLMs. However, incorporating negations in medical contrastive settings can exacerbate semantic overlap, introducing significant noise into the training pipeline.

\subsection{Soft Alignment for CLIP} Although CLIP shows resilience to imbalanced and long-tailed data distributions~\cite{generalization}, its performance deteriorates with highly imbalanced datasets. Re-weighting strategies and specialized loss functions have been explored~\cite{imbalance}, but often lack adaptability to the medical domain. PyramidCLIP~\cite{pyramidclip} and SoftCLIP~\cite{softclip} relax the strict one-to-one constraint of CLIP’s contrastive loss by implementing soft cross-modal alignment based on intra-modal self-similarity. However, this can introduce ambiguity in text embeddings within the clinical domain, leading to noisy outputs and unstable training. Medical datasets, characterized by extreme imbalance demand innovative solutions tailored to medical imaging to effectively address CLIP's limitations.

\begin{figure*}
  \centering
  \includegraphics[width=1.0\linewidth]{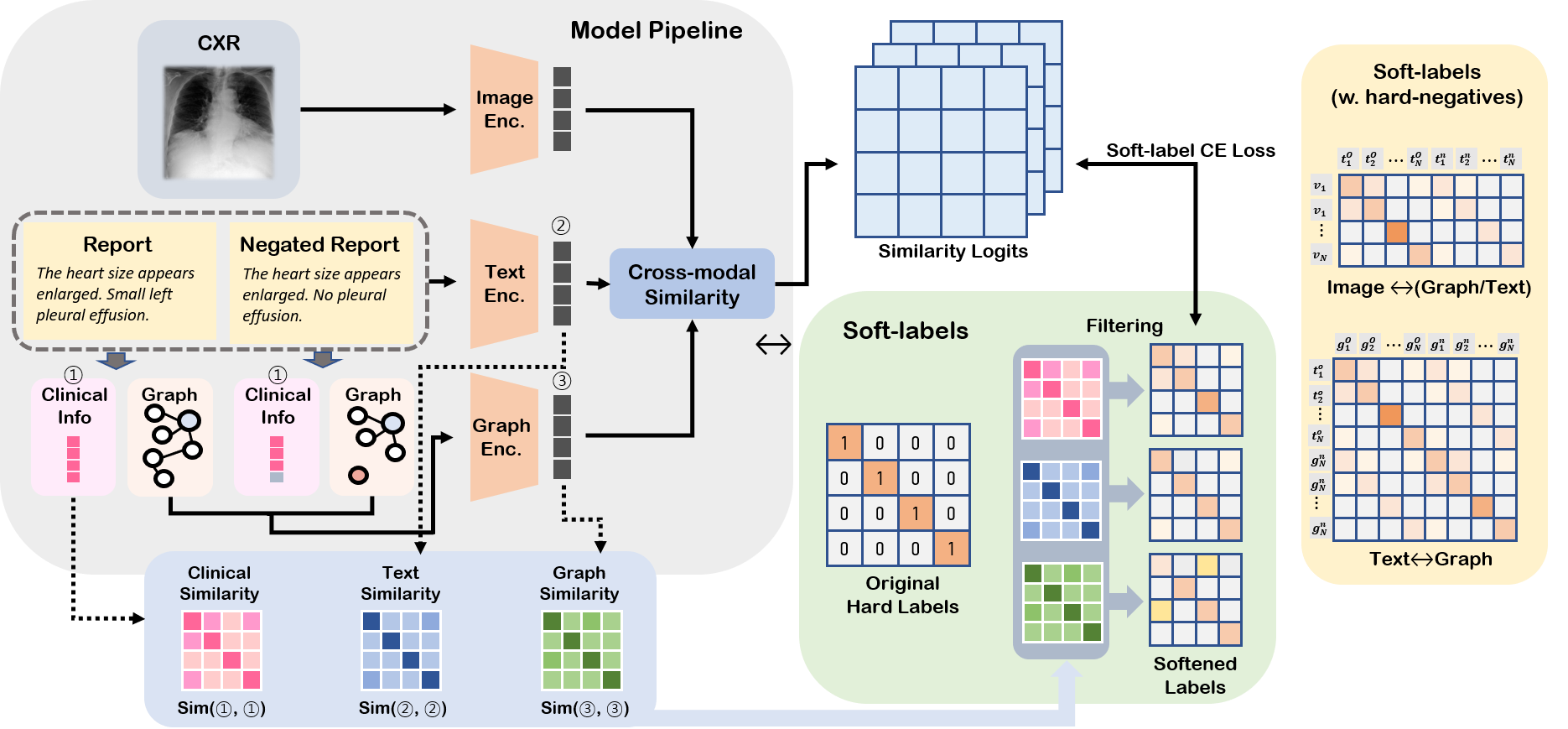}
  \caption{Overview of the proposed pipeline. Hard negative reports are created that differ from the original by only one clinical entity. Embeddings of each modality (CXR, report, graph) are extracted by their encoders, along with clinical labels from the report. Intra-modal self-similarities are computed for clinical labels, text embeddings, and graph embeddings, used as soft labels for each stream. The conventional InfoNCE loss is replaced by KL-Divergence when incorporating softened targets, ensuring labels reflect the textual, clinical, and graphical meanings correctly.}
  \label{fig:pipeline}
\end{figure*}

%% file: sec/3_cxralign.tex
\section{CXR-Align: A Benchmark for CXR-Report Alignment with Negations}
\label{sec:cxrneg}

Negations are rarely present in image-text pairs within general-domain datasets, limiting CLIP models' ability to accurately interpret negated information. In contrast, medical datasets frequently contain negations, which are critical for precise clinical interpretation. To address this gap, we introduce \textit{CXR-Align}, the first benchmark specifically designed to evaluate models’ comprehension of negations in CXR reports—an essential aspect for clinical applications.
\textit{CXR-Align} is synthesized from the test sets of MIMIC~\cite{mimic3}, and OpenI~\cite{openi}. We begin by transforming the original reports using a large language model (LLM), specifically Gemini-Flash~\cite{gemini}, to ensure that each sentence is limited to a single clinical entity~\cite{maira2}. To focus on diagnostically relevant cases, normal CXRs are excluded from this benchmark. We further standardize the reports by removing any temporal references, centering each on a single CXR instance. Using a CXR report labeler, specifically CheXbert~\cite{chexbert}, we identify positive findings, diseases, or medical devices within each report, then randomly select one entity for further processing. We create report variations as follows:
\begin{itemize}
    \item \textbf{Removing positive entity (\(r^r\))}: The LLM removes the selected positive entity from the report, resulting in \(r^r\).
    \item \textbf{Adding negations (\(r^n\))}: From \(r^r\), we generate \(r^n\) by inserting a predefined negative statement about the selected entity.
\end{itemize}

The final dataset contains two triplet structures: \((I, r, r^n)\) and \((I, r, r^r)\). The first triplet assesses the model's ability to understand and correctly reject an incorrectly negated statement, while the second evaluates its grasp of full clinical semantics by identifying the most complete report. This setup enables us to benchmark our model alongside other state-of-the-art vision-language models, focusing on negation comprehension and CXR-report alignment. More details for this dataset can be found on \cref{sec:cxrneg2}.

%% file: sec/4_method.tex
\section{Method}
Our approach to training CLIP models on medical datasets addresses semantic overlap and negation by introducing a clinically-enhanced dynamic soft-label strategy combined with negation-based hard negatives. This section outlines our method for generating negated data as hard negatives, creating dynamic soft labels, and formulating the training loss, enhanced with graph embeddings.

\subsection{Generating Data with Negations for Hard Negative Training}
To improve the model's understanding of negations, we generate hard-negative samples. Following the process from \cref{sec:cxrneg}, we create hard negatives \(r^n\) for abnormal CXRs by introducing negations into the report. For normal CXRs, we randomly select reports containing only a single positive entity, using these as hard negatives rather than generating negated reports. This approach ensures that the hard negative reports differ from the original reports by only one entity, making them challenging for the model to distinguish.

\subsection{Dynamic Soft Contrastive Loss}
To address imbalance and semantic overlap within a batch, we implement dynamic soft labels that reflect clinical similarities between text embeddings and clinical labels. Let \(T_1 \in \mathbb{R}^{B \times D_t}\) denote the L2-normalized text embeddings and \(T_2 \in \mathbb{R}^{B \times D_t}\) the hard negative text embeddings, where \(B\) is batch size, and \(D_t\) is the dimension of the text embeddings. Similarly, \(C_1 \in \mathbb{R}^{B \times D_c}\) and \(C_2 \in \mathbb{R}^{B \times D_c}\) represent the L2-normalized clinical labels and hard negative clinical labels, respectively, with \(D_c\) being the dimension of the clinical labels. We concatenate \(T_1\) and \(T_2\) to form combined text embeddings \(T = [T_1; T_2] \in \mathbb{R}^{2B \times D_t}\) and likewise for the clinical embeddings \(C = [C_1; C_2] \in \mathbb{R}^{2B \times D_c}\). Since embeddings alone may not fully capture clinical semantics~\cite{rexamine}, we use 14 labels extracted from reports by CheXbert~\cite{chexbert} as an alternative source of clinical information.

As in InfoNCE~\cite{infonce} for cross-modal alignment, the normalized cross-modal logits are calculated as:
\begin{align}
    p_{ij}(I, T) &= \frac{\exp(\text{sim}(v_i, t_j)/\tau)}{\sum_{j=1}^{2B} \exp(\text{sim}(v_i, t_j)/\tau)}, \\
    p_{ij}(T_1, I) &= \frac{\exp(\text{sim}(t_i, v_j)/\tau)}{\sum_{j=1}^B \exp(\text{sim}(t_i, v_j)/\tau)},
    \label{eq:infonce}
\end{align}
where \(v \in \mathbb{R}^{B \times D_{img}}\) are the image embeddings, \(\tau\) is the temperature parameter, and \(\text{sim}(\cdot)\) denotes dot-product similarity.

Next, we compute similarity matrices for text and clinical labels:
\begin{equation}
    S_t = T \cdot T^T, \quad S_c = C \cdot C^T,
    \label{modality_sim}
\end{equation}
where \(S_t \in \mathbb{R}^{2B \times 2B}\) and \(S_c \in \mathbb{R}^{2B \times 2B}\) represent intra-modal similarities for text and clinical embeddings, respectively.

Dynamic soft labels are generated by applying thresholds \(\tau_t\) and \(\tau_c\) to retain values above each threshold:
\begin{equation}
    y_t[i, j] = \begin{cases}
        \frac{S_t[i, j] - \tau_t}{1 - \tau_t}, & \text{if } S_t[i, j] > \tau_t, \\
        0, & \text{otherwise},
    \end{cases}
\end{equation}
\begin{equation}
    y_c[i, j] = \begin{cases}
        \frac{S_c[i, j] - \tau_c}{1 - \tau_c}, & \text{if } S_c[i, j] > \tau_c, \\
        0, & \text{otherwise},
    \end{cases}
\end{equation}
where \(y_t \in \mathbb{R}^{2B \times 2B}\) and \(y_c \in \mathbb{R}^{2B \times 2B}\) are then normalized across each row to \(\hat{y}_t\) and  \(\hat{y}_c\) ensuring that the sum of each row equals 1. Note that thresholding is crucial since sharing labels for data with minimal similarity introduces noise.

The image-to-text loss for each similarity measure incorporates KL-Divergence with the generated soft labels and is defined as follows:
\begin{equation}
    \mathcal{L}_{m}(I,T) = \frac{1}{B} \sum_{i=1}^B \text{KL}(\hat{y}_m[i] \parallel p_i(I, T)),
\end{equation}
\begin{equation}
    \mathcal{L}_{c}(I,T) = \frac{1}{B} \sum_{i=1}^B \text{KL}(\hat{y}_c[i] \parallel p_i(I, T)),
    \label{eq:kl}
\end{equation}

where \(\mathcal{L}_{t}(T_1,I)\), \(\mathcal{L}_{c}(T_1,I)\) is also computed in a similar manner.

\subsection{Dynamic Contrastive Loss with Graph Embeddings}
To capture additional clinical relationships, we integrate graph embeddings. Clinical embeddings may lack specific attributes such as location, severity, or size of entities, as they only encode presence. We use RadGraph~\cite{radgraph} to extract graphs from each report, embedding each node with ClinicalBERT~\cite{clinicalbert}. A two-layer Graph Convolutional Network~\cite{gcn} then produces graph embeddings \(G_1 \in \mathbb{R}^{B \times D_t}\) and hard negative graph embeddings \(G_2 \in \mathbb{R}^{B \times D_t}\), which are concatenated as \(G = [G_1; G_2] \in \mathbb{R}^{2B \times D_G}\).
Using graph embeddings \(G\), we compute pairwise similarity:
\begin{equation}
    S_g = G \cdot G^T,
\end{equation}
where \(S_g\) represents graph similarity within the batch. Graph-based soft labels \(y_g\) are generated with a threshold \(\tau_g\):
\begin{equation}
    y_g[i, j] = \begin{cases}
        \frac{S_g[i, j] - \tau_g}{1 - \tau_g}, & \text{if } S_g[i, j] > \tau_g, \\
        0, & \text{otherwise}.
    \end{cases}
\end{equation}
The normalized \(\hat{y}_g\) serves as soft labels, and KL-Divergence loss terms are computed with the graph-related logits extracted as like in \cref{eq:infonce} with \cref{eq:kl} across text, clinical, and graph similarity stream.

The final training loss integrates all cross-modal components as follows:
\begin{align}
    \mathcal{L}_{\text{total}} = \sum_{i=1, i \neq j}^3 \sum_{j=1}^3 \Big( w_T \cdot \mathcal{L}_{t}(M_i, M_j) \nonumber
    & + w_C \cdot \mathcal{L}_{c}(M_i, M_j) \\ + w_G \cdot \mathcal{L}_{g}(M_i, M_j) \Big).
\label{eq:final}
\end{align}

Here, \(M_1\), \(M_2\), \(M_3\) correspond to the image, text, and graph modalities, respectively, where we  use \(T_1\) and \(G_1\) instead of \(T\) and \(G\) when compared with \(I\). \(w_T\), \(w_C\), and \(w_G\) are weighting coefficients for each loss component.

%% file: sec/5_experiment.tex
\section{Experiments \& Results}
Beyond traditional CXR evaluation tasks, including zero-shot, fine-tuned classification and report retrieval, we introduce new novel tasks such as the \textit{CXR-Align} benchmark, RSNA-Abnormal (RSNA-\textit{ab}) classification, adversarial prediction, and normal case detection to further assess our model’s clinical understanding and robustness.

\begin{table*}[h]
    \centering
    \setlength{\tabcolsep}{2pt} 
    \small
    \resizebox{\textwidth}{!}{
    \begin{tabular*}{\textwidth}{@{\extracolsep{\fill}}l ccc ccc ccc c c c @{}}
        \toprule
        & \multicolumn{3}{c}{RSNA} & \multicolumn{3}{c}{RSNA-\textit{ab}} & \multicolumn{3}{c}{SIIM} & VinDR & Chexpert & CXR14 \\
        \cmidrule(lr){2-4} \cmidrule(lr){5-7} \cmidrule(lr){8-10} \cmidrule(lr){11-11} \cmidrule(lr){12-12} \cmidrule(lr){13-13}
        Model & ZS & FT10 & FT100 & ZS & FT10 & FT100 & ZS & FT10 & FT100 & ZS & ZS & ZS \\
        \midrule
        ConVIRT~\cite{convirt} & 75.6 & 78.1 & 80.3 & 68.2 & 75.2 & 76.7 & 68.3 & 78.9 & 81.4 & 68.6 & 39.4 & 51.2 \\
        BIOVIL~\cite{biovil}  & 84.3 & 87.6 & 89.1 & 73.8 & 82.0 & 83.0 & 78.6 & 86.0 & 87.3 & 77.2 & 41.5 & 55.4 \\
        BIOVIL-T~\cite{biovilt}   & \textbf{87.8} & 88.2 & 89.2 & \textbf{79.4} & 82.1 & 83.3 & 74.9 & 86.8 & 87.9 & 77.4 & 44.2 & 53.7 \\
        CXRCLIP~\cite{cxrclip} & 81.4 & 88.1 & 89.3 & 72.0 & 83.6 & 84.0 & 85.4 & 87.2 & 88.7 & 78.3 & 53.0 & 55.9 \\
        \midrule
        CLIP    & 81.2 & 88.3 & 89.1 & 70.6 & 83.7 & 84.0 & 74.3 & 87.8 & 88.0 & 76.1 & 52.3 & 56.7 \\
        SOFTCLIP~\cite{softclip}  & 76.6 & 79.1 & 81.1 & 67.8 & 73.1 & 76.2 & 70.1 & 78.3 & 80.3 & 73.1 & 47.1 & 54.2 \\
        CLIP-D\(_t\)  & 81.6 & 88.5 & 89.5 & 71.8 & 83.8 & 84.2 & 80.1 & 88.0 & 88.3 & 78.1 & 54.1 & 59.2 \\
        CLIP-D\(_{t+c}\)  & 84.4 & 89.2 & 90.0 & 74.5 & 84.2 & 84.8 & 84.8 & 88.5 & 88.6 & 78.2 & 54.1 & 61.2 \\
        CLIP\(^{N}\)  & 82.0 & 88.2 & 89.0 & 72.5 & 83.3 & 83.9 & 74.1 & 88.4 & 88.9 & 76.1 & 53.7 & 57.2 \\
        CLIP\(^{N}\)-D\(_{t+c}\) & 86.4 & \textbf{90.7} & \textbf{91.2} & 78.1 & 84.5 & 85.3 & 85.6 & 89.2 & 89.8 & \textbf{79.1} & 54.4 & 62.8 \\
        CLIP\(^{G}\)  & 81.8 & 88.6 & 89.2 & 70.5 & 83.3 & 84.1 & 70.8 & 87.3 & 87.8 & 77.0 & 52.6 & 57.1 \\
        CLIP\(^{G}\)-D\(_{t+c+g}\) & 85.1 & 89.9 & 90.5 & 74.9 & 84.2 & 85.1 & 84.5 & 88.7 & 89.0 & 78.3 & 56.1 & 62.3 \\
        CLIP\(^{N, G}\)-D\(_{t+c+g}\) & 86.6 & \textbf{90.7} & 91.1 & 78.3 & \textbf{84.8} & \textbf{85.4} & \textbf{87.2} & \textbf{89.6} & \textbf{90.2} & 78.8 & \textbf{57.3} & \textbf{63.0} \\
        \bottomrule
    \end{tabular*}
    }
    \caption{Performance comparison across datasets for zero-shot (ZS) and fine-tuned (FT) entity classification for models trained on MIMIC. FT10 and FT100 denote fine-tuning with 10\% and 100\% of the data, respectively. The upper part shows SOTA models' performance; the lower part shows performance improvements as features are added from the baseline. Here, \(N\) denotes training with hard negatives, \(G\) denotes training with graph embeddings and \(D\) indicates training with dynamic soft labels based on  \(t\) for textual similarity, \(c\) for clinical similarity, and \(g\) for graph similarity. AUC is measured for RSNA and SIIM dataset while accuracy is measured for the others.}
    \label{tab:entity_classification}
\end{table*}

\subsection{Dataset}
For training, we use the MIMIC dataset, where original reports are split and prior references are omitted as described in \cref{sec:cxrneg}. All datasets undergo our preprocessing pipeline detailed in \cref{sec:preprocess}. Additionally, we utilize a private tertiary hospital dataset spanning 20 years with the last year as test set for our novel normal case detection task.

Evaluation is conducted on multiple datasets: zero-shot and fine-tuned classification on RSNA-Pneumonia, RSNA-\textit{ab} (a subset of RSNA where the model is required to distinguish pneumonia cases from all other abnormal cases), SIIM Pneumothorax, VinDr~\cite{nguyen2022vindr}, CXR14~\cite{cxr14}, and Chexpert; adversarial prediction on CXR14 following the approach in Probmed~\cite{probmed}; and normal case detection on OPEN-I dataset. The CXR-Align benchmark and report retrieval task is evaluated with OpenI, MIMIC, and Chexpert datasets. Note that RSNA, VinDr, SIIM, and Chexpert test sets align with those used in GLORIA~\cite{huang2021gloria}. For CXR14, we use the data selected by Probmed.

\subsection{Model Settings}
We utilize a Swin-Tiny~\cite{swin} model as the image encoder, BioClinical-BERT~\cite{clinicalbert} as the text encoder, and a 2-layer GCNN for graph encoding. Input resolution is set to $224 \times 224$ pixels. Additional details are provided in \cref{sec:model_info}.

\begin{table*}[h]
    \centering
    \small
    \setlength{\tabcolsep}{2pt}
    \begin{tabular*}{\textwidth}{@{\extracolsep{\fill}}l c c c c c c c@{}}
        \toprule
        CXR Model & \multicolumn{2}{c}{RSNA} & \multicolumn{2}{c}{RSNA-\textit{ab}}& \multicolumn{2}{c}{SIIM}&\multicolumn{1}{c}{NCD}\\
        \cmidrule(lr){2-3} \cmidrule(lr){4-5} \cmidrule(lr){6-7} \cmidrule(lr){8-8}
        & ZS & FT10 & ZS & FT10 & ZS & FT10 & ACC\\
        \midrule
        CLIP & 77.2 \textcolor{red}{(-4.0)} & 87.6 \textcolor{red}{(-0.7)} & 65.0 \textcolor{red}{(-5.6)} & 82.9 \textcolor{red}{(-0.8)} & 72.4 \textcolor{red}{(-1.9)} & 87.5 \textcolor{red}{(-0.3)} & 84.3 \\
        CLIP-D\(_t\) & 78.8 \textcolor{red}{(-2.8)} & 88.5 (-0.0) & 68.8 \textcolor{red}{(-3.0)} & 83.7 \textcolor{red}{(-0.1)} & 79.3 \textcolor{red}{(-0.8)} & 87.9 \textcolor{red}{(-0.1)} & 81.4 \\
        CLIP-D\(_{t+c}\) & 82.1 \textcolor{red}{(-2.3)} & 89.4 \textcolor{blue}{(+0.2)} & 72.4 \textcolor{red}{(-2.1)} & 84.2 (+0.0) & 84.1 \textcolor{red}{(-0.7)} & 88.9 \textcolor{blue}{(+0.4)} & 86.8\\
        CLIP\(^{G}\)-D\(_{t+c+g}\) & 83.2 \textcolor{red}{(-1.9)} & 90.1 \textcolor{blue}{(+0.2)} & 72.8 \textcolor{red}{(-2.1)} & 84.5 \textcolor{blue}{(+0.3)} & 83.8 \textcolor{red}{(-0.7)} & 88.8 \textcolor{blue}{(+0.1)} & 85.6 \\
        \bottomrule
    \end{tabular*}
    \caption{Ablation study for the dynamic soft labels in a manually set up imbalanced dataset. Performance differences are measured compared to using only the MIMIC dataset for training, as in \cref{tab:entity_classification}. Normal Case Detection (NCD) is performed with OpenI normal CXRs, where the model is required to retrieve one normal report from 2,999 abnormal reports.}
    \label{tab:performance_comparison}
\end{table*}

\subsection{Classification}
\label{ex:main}
In this section, we demonstrate that entity classification tasks benefit significantly from all proposed methodologies. \cref{tab:entity_classification} shows the zero-shot and fine-tuned performance across tasks, methods, and training datasets.

\textbf{Dynamic Soft Labeling}: Utilizing soft labels based on text similarity alone provides performance improvements over the baseline CLIP. Incorporating clinical and relational similarities further enhances performance, leading to even better results. Our approach outperforms other SOTA methods on most benchmark datasets without requiring lateral images, external knowledge, masked modeling, or MVS~\cite{declip} methods. The performance enhancement is particularly evident on the SIIM dataset as more similarity measures are incorporated into the dynamic soft label approach. Although our final model's zero-shot performance is slightly below that of BIOVIL-T, our fine-tuned performance is superior, suggesting that our method extracts richer representations reflecting clinical information. Notably, our soft labels lead to more stable and improved performance compared to SOFTCLIP. Instead of relying solely on text similarity, we employ thresholding and distribute role among text, clinical, and relational graph similarities, helping the model allocate its outputs appropriately across categories and enhancing clinical comprehension. A hypothesis of this effect is discussed on \cref{sec:motive}.

\textbf{Hard Negatives}: Using negations as hard negatives does not improve zero-shot and fine-tuned performance when used alone, possibly due to increased overlap of clinical semantics within the batch. However, employing the dynamic soft label approach effectively addresses this issue, stabilizing the effect of hard negatives and boosting both zero-shot and fine-tuned performances. As we incorporate each soft-label approach, performance improves across benchmarks, especially in fine-tuned classification.

\textbf{Graph Embeddings}: Incorporating graph contrastive loss using graph embeddings with dynamic soft labels further enhances performance, and combining this with hard negatives leads to notable gains. Overall, this method synergies well with both dynamic soft labels and hard negatives.

\subsection{Adversarial Prediction}
\label{ex:adverserial}
Following the approach of ProbMed~\cite{probmed}, we constructed a zero-shot task comprising a positive first query and a negative second query, where the model should correctly identify the positive entity in the first query and the negative entity in the second query in sequence. Although the original paper evaluated Large Language Models (LLMs), we adapted this task to evaluate CLIP. This is a complex task requiring the model to fully understand the image and recognize which entities are present and which are absent. As shown in \cref{tab:adversarial_cls}, while all of the SOTA models performed below chance level, our model achieved significantly better results compared to both chance level and the best performing model by a significant margin.

\begin{table}[h]
    \centering
    \small
    \begin{tabular}{lc}
        \toprule
        Model & Adversarial CLS \\
        \midrule
        CXRCLIP & 21.4 \\
        BIOVIL  & 23.3 \\
        BIOVIL-T   & 14.0 \\
        MedCLIP & 11.5 \\
        GLORIA  & 12.0 \\
        OURS    & \textbf{34.4} \\
        \bottomrule
    \end{tabular}
    \caption{Adversarial prediction accuracy where the model is required to guess both positive and negative entities correctly in a zero-shot setting.}
    \label{tab:adversarial_cls}
\end{table}

\begin{table*}[h]
    \centering
    \small
    \begin{tabular}{l cccc cccc cccc cccc}
        \toprule
        & \multicolumn{4}{c}{MIMIC} & \multicolumn{4}{c}{Chexpert} & \multicolumn{4}{c}{Open-I} \\
        \cmidrule(lr){2-5} \cmidrule(lr){6-9} \cmidrule(lr){10-13}
        Model & @5 & F1 & Recall & Precision & @5 & F1 & Recall & Precision & @5 & F1 & Recall & Precision  \\
        \midrule
        MedCLIP  & 1.3 & 15.9 & 10.7 & 24.2 & 2.8 & 3.8 & 2.8 & 4.4 & 0.4 & 2.0 & 1.4 & 3.2 \\
        BIOVIL   & 10.9 & 36.4 & 35.6 & 38.9 & 10.5 & 24.6 & 24.9 & 25.6 & 3.2 & 20.6 & 22.9 & 22.3 \\
        BIOVIL-T  & 13.3 & 36.7 & 36.4 & 39.1 & 11.1 & 24.4 & 23.9 & 28.5 & 3.7 & 21.1 & 25.1 & 22.7 \\
        \midrule
        CLIP     & 35.5 & 44.8 & 45.0 & 44.6 & 24.5 & 35.4 & 33.3 & 39.9 & 6.4 & 28.7 & 30.6 & 28.5 \\
        CLIP\(^{N}\)-D\(_{t+c}\)   & 33.9 & 46.0 & 49.7 & 43.9 & 27.9 & 38.0 & 39.3 & 30.0 & 6.2 & 28.8 & \textbf{33.5} & 24.6 \\
        CLIP\(^{N, G}\)-D\(_{t+c+g}\)  & \textbf{38.8} & \textbf{50.6} & \textbf{51.5} & \textbf{50.3} & \textbf{29.3} & \textbf{42.0} & \textbf{43.2} & \textbf{36.6} & \textbf{7.8} & \textbf{29.0} & 31.4 & \textbf{29.2} \\
        \bottomrule
    \end{tabular}
    \caption{Retrieval performance on MIMIC, CheXpert, and OpenI, evaluated using Top-5 accuracy (@5) and CheXbert-based F1, Recall, and Precision scores.}
    \label{tab:retrieval_performance}
\end{table*}

\subsection{Ablation with Imbalanced Dataset and Normal Case Detection}
\label{ex:normalthinning}
To simulate the effects of training in a general hospital setting, where class imbalance is significant, we added 130,000 normal CXR cases (all labeled as "No active lung lesion") from our private dataset to the MIMIC training data. This addition introduces substantial imbalance, with normal CXRs constituting over half of the dataset and a large number of duplicate reports. As shown in \cref{tab:performance_comparison}, this imbalance reduces zero-shot accuracy. However, our dynamic soft label approach mitigates this performance drop, and when clinical similarities are applied, fine tuned performance surpasses that of the original model.

This raises a key question: Why add more normal data and increase imbalance rather than remove excess normal data for a balanced dataset?" The answer lies in the normal case detection task, which assesses the model's ability to identify normal cases. This challenging task requires the model to retrieve the one normal CXR report from among 2,999 abnormal reports in the test set—a \textit{needle-in-a-haystack} scenario. Accuracy is measured by whether the model successfully retrieves the normal report.

In evaluating these measures with CXR-CLIP and our CLIP\(^{N,G}\)-D\(_{t+c+g}\) in \cref{tab:entity_classification}, we observe low accuracy (0.7, 3.1) percent, respectively. However, including normal data in the training set raises these measures to over 80\%, demonstrating the importance that inclusion of normal data is crucial for enhancing the model's comprehension of normal cases.

\begin{table}[h]
    \centering
    \small
    \begin{tabular}{l ccc ccc}
        \toprule
        & \multicolumn{2}{c}{MIMIC} & \multicolumn{2}{c}{Open-I} \\
        \cmidrule(lr){2-3} \cmidrule(lr){4-5}
        Model & A & B & A & B \\
        \midrule
        GLORIA  & 50.0 & 34.4 & 59.6 & 35.0 \\
        BIOVIL  & 60.5 & 61.0 & 60.5 & 58.3 \\
        BIOVIL-T   & 64.3 & 65.1 & 60.9 & 62.6 \\
        CXRCLIP & 78.3 & 73.6 & 72.2 & 68.7 \\
        \midrule
        CLIP    & 75.4 & 72.4 & 62.7 & 62.6 \\
        CLIP\(^{N}\)  & \textbf{97.3} & 71.7 & \textbf{96.7} & 62.3 \\ 
        OURS & 96.5 & \textbf{80.1} & 96.4 & \textbf{73.8} \\ 
        \bottomrule
    \end{tabular}
    \caption{CXR-Align performance across different datasets. The model is required to perform two tasks: (A) selecting between the original report \(r\) and a report \(r^n\) where an entity present in the CXR has been negated; (B) select between the original report \(r\) and a report \(r^r\) where a sentence related to a specific entity has been removed.}
    \label{tab:cxr_neg_performance}
\end{table}

\subsection{CXR-Align Benchmark Evaluation}
\label{ex:cxralign}
The evaluation on \textit{CXR-Align} demonstrates that introducing hard negatives enhances our model's understanding of negation. As shown in \cref{tab:cxr_neg_performance}, our model significantly outperforms other SOTA models. Notably, the performance of our model trained with hard-negatives on the negation-related task is unexpectedly high, leading us to hypothesize that the model may be learning to avoid unnatural negations by exploiting shortcuts.

To address this issue, we conducted a second task where the generated negated sentence was omitted(\(r^r\)) without being replaced with negations. Our final model also showed improved performance on this task, suggesting that introducing negations can enhance the full alignment between CXR images and reports. It is important to note that using negation-based hard negatives alone does improve performance on task 1, but the performance drops on task 2 compared to the baseline, indicating that semantic overlap may have introduced noise that hinders the model's ability to learn clinical concepts.

\subsection{Report Retrieval}
As shown in \cref{tab:retrieval_performance}, our model achieves competitive retrieval performance compared to other state-of-the-art models, with a particularly strong showing on the CheXbert F1 score. This implies that our model captures more clinically meaningful features from the reports and forms a better alignment compared to other methods.

%% file: sec/6_discussion.tex
\section{Discussion}
Our work presents a method for extracting medical-focused representations that addresses key challenges in adapting general-domain models to the medical domain, specifically semantic overlap (mostly caused by data imbalance), and negation handling. Our approach consistently outperforms both baseline and state-of-the-art models across classical tasks and novel benchmarks, demonstrating robustness and effectiveness. Key insights from our findings include: (1) While using negations as hard negatives alone provided limited benefits, combining them with dynamic soft labels significantly improved performance (\cref{ex:main}); (2) our methods enhanced comprehensive clinical understanding, improving performance in adversarial tasks and alignment benchmarks (\cref{ex:cxralign}, \cref{ex:adverserial}); and (3) the inclusion of normal and duplicate reports contributed positively to training on imbalanced datasets, proving valuable insights for general hospital data where data imbalances are common (\cref{ex:normalthinning}).

A potential area for refinement lies in refining the RadGraph~\cite{radgraph}-based graph representation by focusing on structured elements such as \textit{location}, \textit{severity/size}, and \textit{entity}, which could yield more precise clinical representations. Additionally, refining the text encoders~\cite{llm2vec, llm2clip}, with a better understanding of compositional context, remains important.

For clinical similarity, we leveraged CheXbert~\cite{chexbert} outputs rather than embeddings, bypassing some limitations associated with embedding-based similarity measures. However, CheXbert does not encompass all clinical entities, and using more comprehensive labels could further enhance model performance. Future improvements might include embeddings that better capture the unique semantics of chest X-rays, thereby deepening the model’s understanding of clinical relationships. Additionally, exploring alternative similarity measures beyond cosine similarity \cite{lai2024carzero} could yield further improvements.

Our method was designed to integrate text, clinical, and graph similarities to capture the complexity of medical image interpretation in global scale. While results show notable improvements, there remains room to refine this approach for even greater impact in clinical applications.

%% file: sec/7_conclusion.tex
\section{Conclusion}
In this work, we addressed two pivotal challenges in medical vision-language processing—data imbalance and negation handling—by introducing a specialized method that bridges the gap between general and medical domains. Our approach employs clinically-enhanced dynamic soft labels to mitigate semantic overlaps, incorporates negation-based hard negatives to improve the model's comprehension of complex clinical semantics, and integrates graph embeddings while leveraging clinical, relational, and textual similarities. This synergy yields substantial improvements over baseline and state-of-the-art models across various benchmarks. The \textit{CXR-Align} benchmark also highlights our model's superior ability to process negations—an often overlooked yet crucial component in medical reporting. Overall, this study paves the way for more effective medical vision-language models that address the unique challenges of clinical environments, advancing the development of reliable AI tools in healthcare.

%% file: sec/X_suppl.tex
\clearpage
\setcounter{page}{1}

\appendix
\renewcommand{\thesection}{\Alph{section}} 

\section{Motivation}

In general-domain datasets, captions involve millions of unique objects, scenes, and entities interacting in a multitude of combinations. Due to the diverse nature of general-domain data, contrastive learning is highly effective, as diversity is guaranteed even with random sampling of data into a batch. However, in medical settings, there are far fewer entities, and their relationships are limited, which does not align well with the objectives of contrastive learning.

\subsection{Imbalance}
Clinical data is often highly skewed, containing many duplicate templated reports, as shown in \cref{tab:skew}. Even when reports differ slightly in wording, semantically identical information still limits the effectiveness of standard contrastive learning. This has led many medical researchers and companies to discard duplicates and train models in a more balanced setting. However, fully normal chest X-rays (CXRs) contain crucial information for triage in clinical practice, as identifying normal cases can significantly reduce radiologists' workload. Our goal, therefore, is to develop a method that leverages all data—including duplicates—without discarding valuable information.

\begin{figure}[t]
    \centering
    \includegraphics[width=1\linewidth]{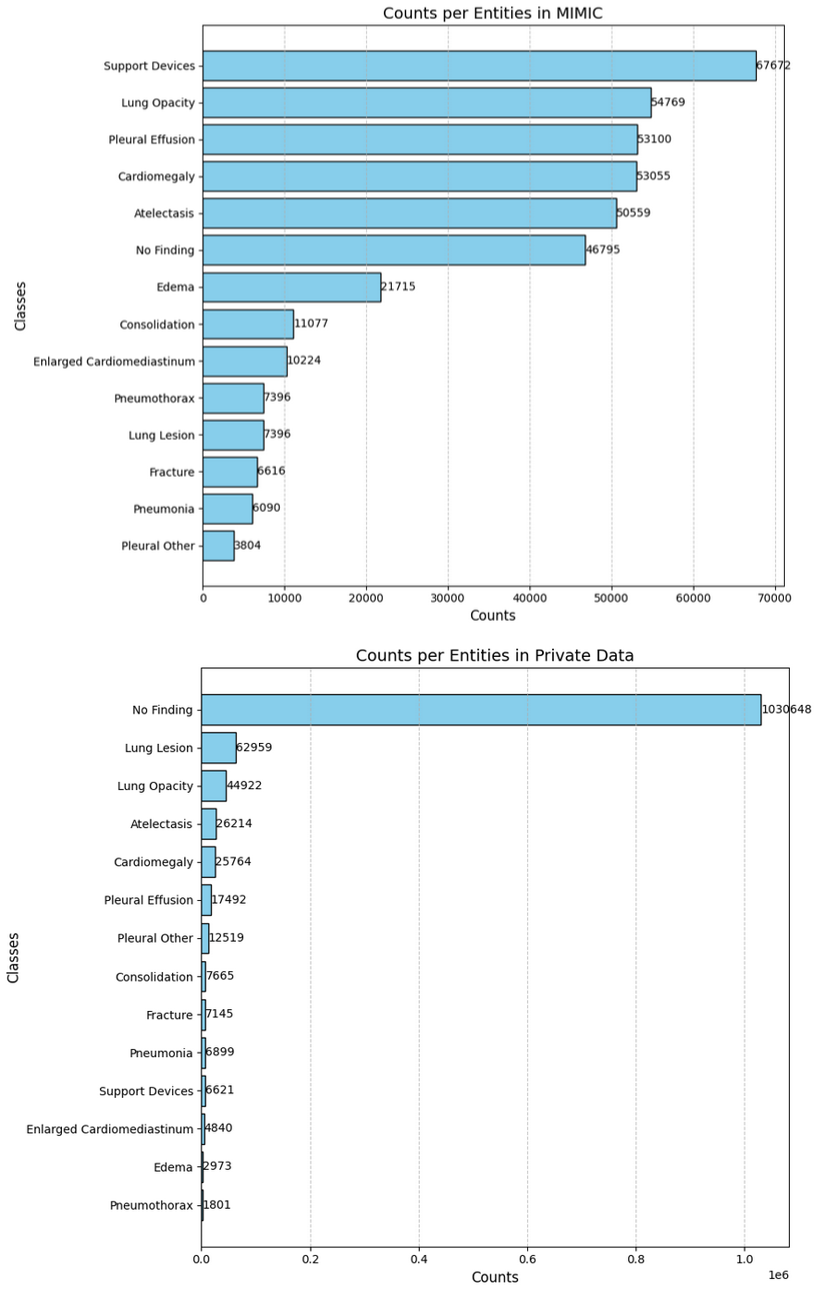}
    \caption{Counts of clinical entities in the whole MIMIC training set and a private dataset collected from a tertiary hospital. The private dataset comprises around 1.3 million records collected over 20 years, each from unique patients.}
    \label{fig:all_entity}
\end{figure}

Furthermore, many reports are semantically similar even if the textual expressions differ. This occurs because there is an imbalance in the entities themselves; similar symptoms are commonly found across medical reports. This can cause semantic overlaps within a batch, where larger batch sizes might introduce more complexity in a contrastive learning context. As illustrated in \cref{fig:all_entity}, when trained with clinical data, one must consider this imbalance of clinical findings within the dataset. Using a general hospital dataset—which has more long-tailed characteristics compared to public data—could introduce noise into the training process due to this imbalance.

\begin{table}[h!]
\centering
\renewcommand{\arraystretch}{1.2} 
\begin{tabular}{|p{0.8\columnwidth}|r|}
\hline
\multicolumn{2}{|c|}{\textbf{Impression}} \\
\hline
No acute cardiopulmonary process. & 37,962 \\
\hline
No acute cardiopulmonary abnormality. & 10,806 \\
\hline
No acute intrathoracic process. & 10,744 \\
\hline
\multicolumn{2}{|c|}{\textbf{Findings}} \\
\hline
Heart size is normal. The mediastinal and hilar contours are normal. The pulmonary vasculature is normal. Lungs are clear. No pleural effusion or pneumothorax is seen. There are no acute osseous abnormalities. & 2,209 \\
\hline
PA and lateral views of the chest provided. There is no focal consolidation, effusion, or pneumothorax. The cardiomediastinal silhouette is normal. Imaged osseous structures are intact. No free air below the right hemidiaphragm is seen. & 1,763 \\
\hline
The lungs are clear without focal consolidation. No pleural effusion or pneumothorax is seen. The cardiac and mediastinal silhouettes are unremarkable. & 1,635 \\
\hline
\end{tabular}
\caption{Most frequent reports from MIMIC impressions and findings. Note that the counts differ from \cref{tab:report_counts} since the reports used in training prioritize findings over impressions.}
\label{tab:skew}
\end{table}

\subsection{Similarity}
\label{sec:motive}
Standard contrastive learning frameworks typically pull positive pairs together and push negative pairs apart. From a clinical perspective, it would be beneficial if similarity could be weighted according to clinical context. For instance, a report noting “Right large pleural effusion. No pneumothorax.” should be considered closer to “Right small pleural effusion.” than to “No pleural effusion. Cardiomegaly exists,” reflecting the clinical relevance of both findings. This is why we incorporate soft labels using similarity measures rather than a uniform distribution of soft labels, which has already been shown to be beneficial in~\cite{softclip}. Notably, using this characteristic, we can also handle duplicates or overlaps of clinical semantics since this method shares labels with similar or identical data within the batch.

\begin{figure*}[t]
    \centering
    \includegraphics[width=1\linewidth]{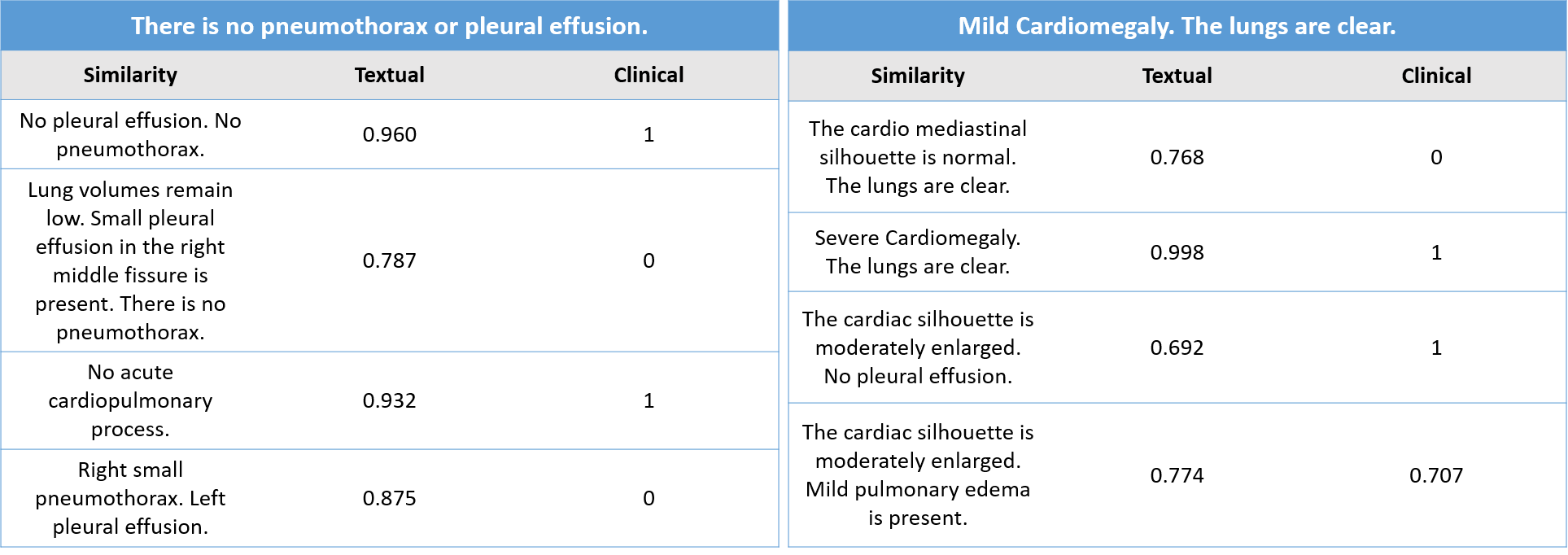}
    \caption{Comparison of textual, clinical similarity between reports.}
    \label{fig:similarity}
\end{figure*}

We explore three types of similarity—textual, clinical, and graph-based—to achieve this nuanced approach. Similarity measures play a crucial role in contrastive learning, particularly in the medical domain. The SOFTCLIP~\cite{softclip} method, which also uses soft labels, relies primarily on textual similarity and is not well-suited for medical data where textual and clinical meanings often diverge. As shown in \cref{fig:similarity}, textual similarity alone does not align well with clinical importance. For example, for the report “Mild cardiomegaly. The lungs are clear,” the textual similarity score is higher with “The cardiomediastinal silhouette is normal. The lungs are clear.” than with “The cardiac silhouette is moderately enlarged. No pleural effusion.” Although the latter is closer in clinical meaning, textual similarity alone fails to capture this. Therefore, using solely textual similarity as soft labels can inadvertently bring unrelated reports closer rather than pushing them apart. This effect is demonstrated in \cref{ex:main} where SOFTCLIP performs worse than the baseline model.

While clinical similarity captures context better than text alone, it does not account for critical details like severity or location, such as “severe” or “mild.” To address this, we introduce graph similarity, which can capture these nuanced attributes and improve alignment.

\begin{figure}[t]
    \centering
    \includegraphics[width=1\linewidth]{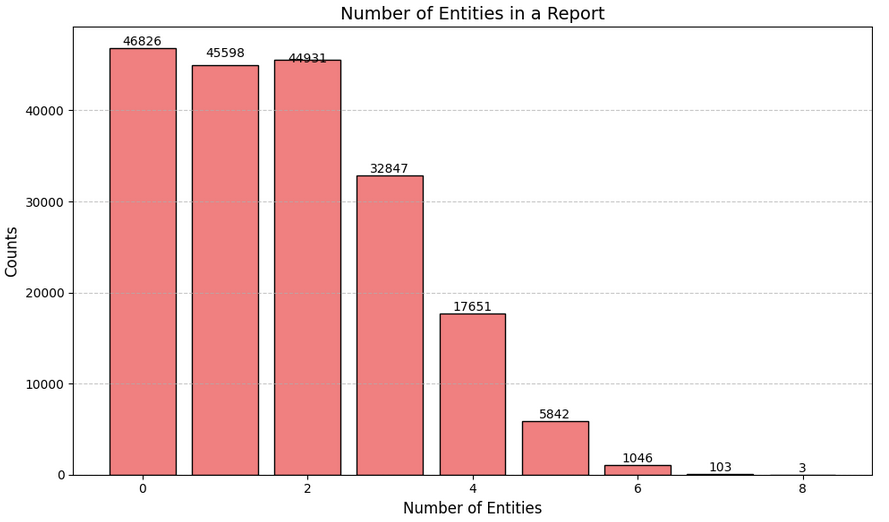}
    \caption{Counts of clinical entities in reports for the MIMIC training set.}
    \label{fig:entity_value_counts}
\end{figure}

\subsection{Negation}
\begin{figure}[t]
    \centering
    \includegraphics[width=1\linewidth]{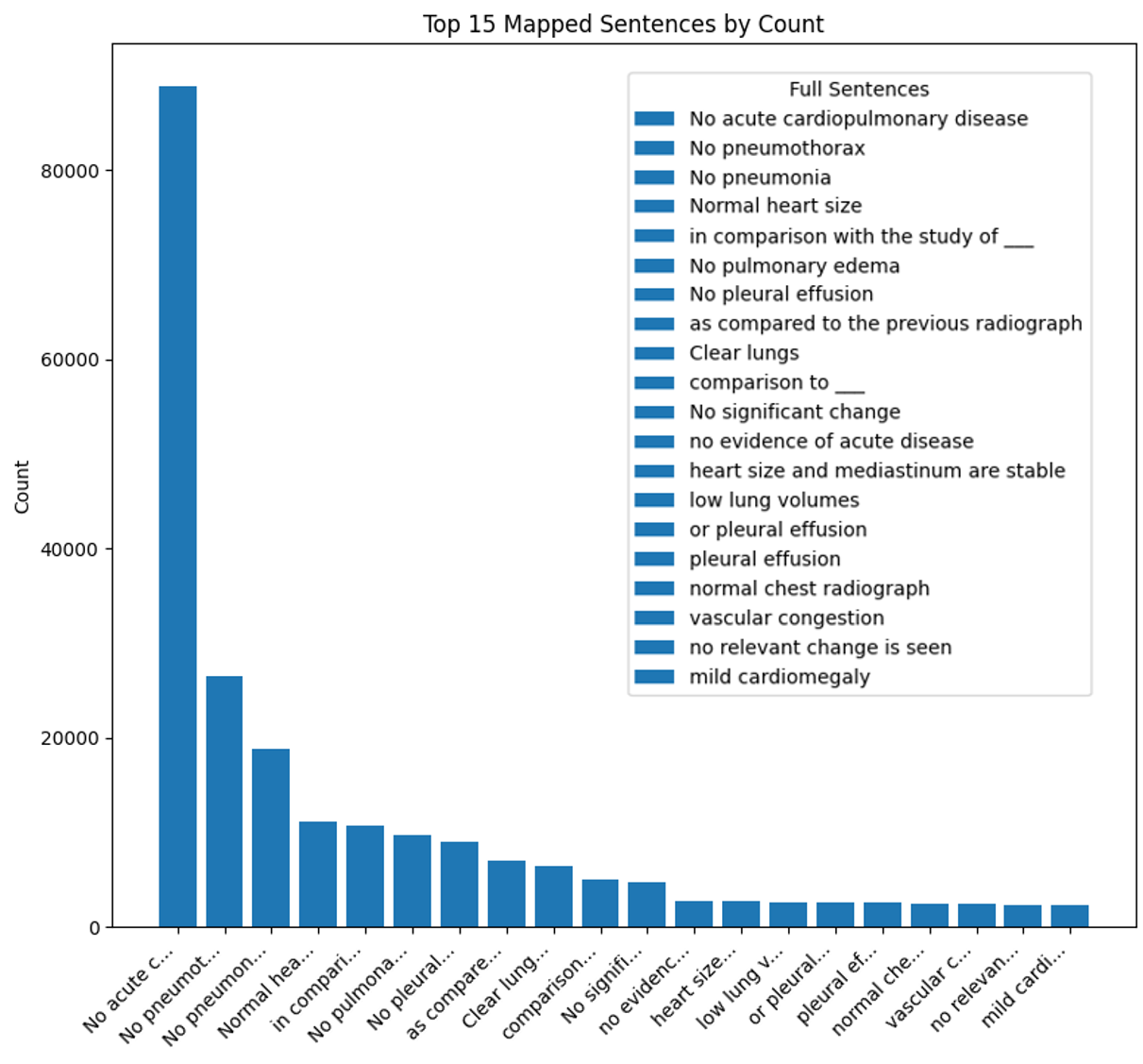}
    \caption{N gram frequent keyword extraction for MIMIC reports. The list is sorted by the top most frequently used phrases.}
    \label{fig:negngram}
\end{figure}

Negations are prevalent in medical reports, as illustrated in \cref{fig:negngram}, where negated terms dominate the dataset. Unlike general domains, medical reports use diverse negation forms, such as “resolved,” “removed,” or “rule out,” in addition to common terms like “no” or “not.” Understanding negation is critical for accurate model performance, but using negated terms as hard negatives in standard contrastive learning often introduces noise. This is why, even though negations are a serious concern, few studies attempt to tackle this issue.

For example, negating the report “Pneumothorax is present on the right upper lung zone” to “No pneumothorax” would yield a hard negative that overlaps semantically with other normal CXRs or cases without pneumothorax in the same batch, causing confusion. As shown in \cref{fig:entity_value_counts}, using negation as a hard negative will introduce more overlaps as entity counts become smaller in the report. By incorporating dynamic soft labels, we can address this issue, allowing the model to handle clinical semantics effectively without adding noise from negated terms.

\section{Dataset}
\label{sup:dataset}
\subsection{Dataset Preprocessing}
\label{sec:preprocess}
\begin{figure}[t]
    \centering
    \includegraphics[width=1\linewidth]{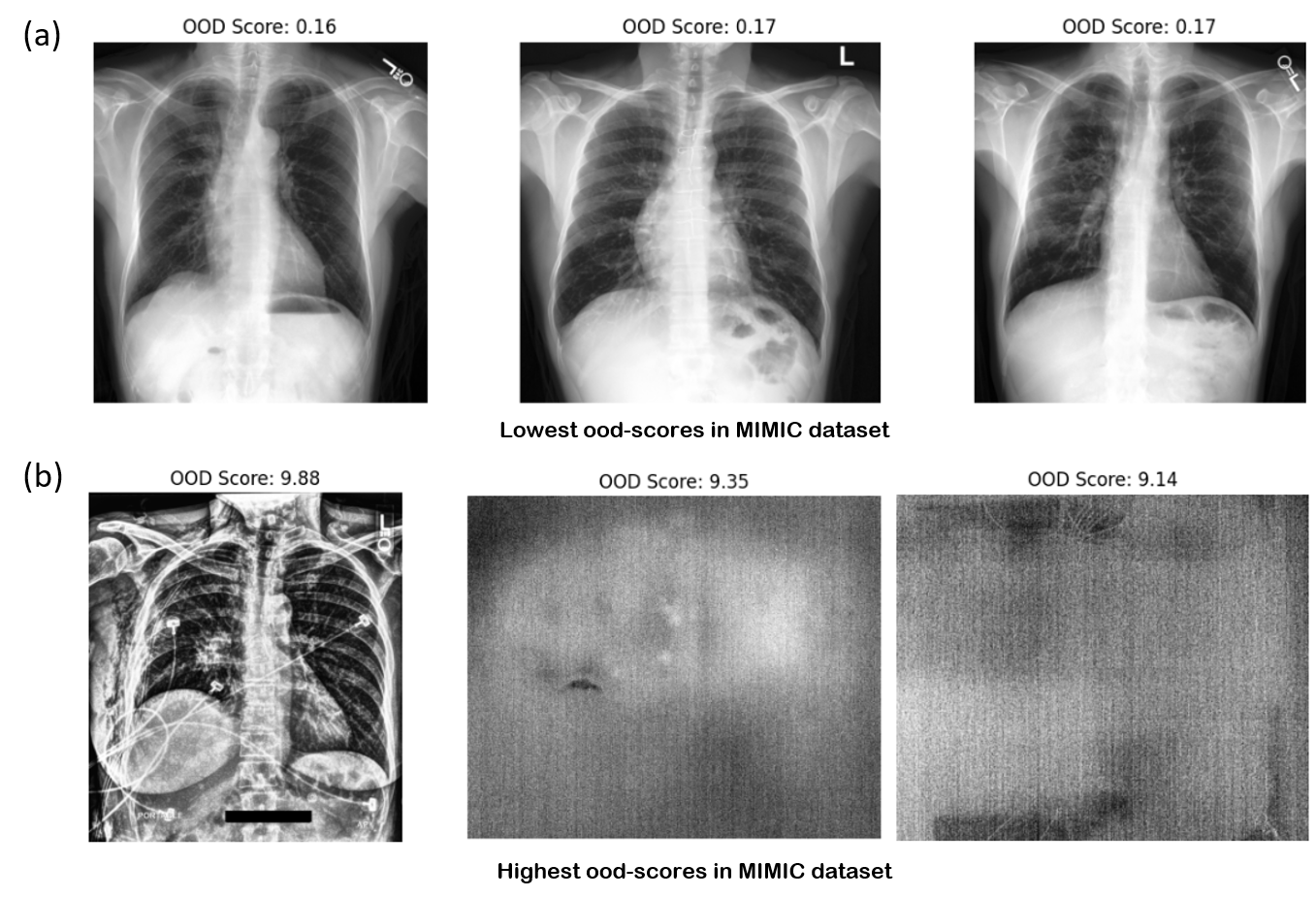}
    \caption{(a) Data with the lowest OOD score in the MIMIC dataset. (b) Data with the highest OOD score in the MIMIC dataset. The OOD detection model is implemented using the DeepMCDD pipeline.}
    \label{fig:ood}
\end{figure}

All CXR images undergo preprocessing through a pipeline that includes monochrome fixation, rotation correction, out-of-distribution (OOD) filtering, and view position selection. The monochrome fixation and rotation correction models were trained on Chexpert dataset using a MobileNetV3 CNN architecture, while view position and OOD detection utilize a DeepMCDD pipeline with a ResNet34 backbone. An example of image post-processing is shown in \cref{fig:ood}. All images are resized to $224 \times 224$ pixels and min-max normalized.

\subsection{Dataset Split}
Details of the training, validation, and test splits for our experiments (\cref{ex:main} and \cref{ex:normalthinning}) are provided in \cref{tab:data_summary_vertical}.  We use the same dataset splits as GLORIA~\cite{huang2021gloria} for CheXpert, VinDR, RSNA, and SIIM, while the CXR14 test set follows the split from ProbMed~\cite{probmed}. For MIMIC and OpenI, we exclude lateral and OOD images to ensure data consistency.

\begin{table}[h]
    \centering
    \begin{tabular}{l ccc}
        \toprule
        Dataset & Train & Valid & Test \\
        \midrule
        MIMIC-CXR   & 194,847 & 1,984 & 2,490 \\
        \midrule
        CheXpert   & - & -   & 1000 \\
        VinDR       & -  & - & 3,000 \\
        RSNA        & 18,678  & 4,003 & 4,003 \\
        RSNA-ab       & -   & - & 3,165 \\
        SIIM        & 8,432   & 1,808 & 1,807 \\
        Open-I      & -       & -     & 3,318 \\
        CXR14       & -       & -     & 880 \\
        \bottomrule
    \end{tabular}
    \caption{Data Summary for training and evaluation.}
    \label{tab:data_summary_vertical}
\end{table}

\subsection{CXR-Align}
\label{sec:cxrneg2}
\subsubsection{Counts}
\begin{figure*}[t]
    \centering
    \includegraphics[width=1\linewidth]{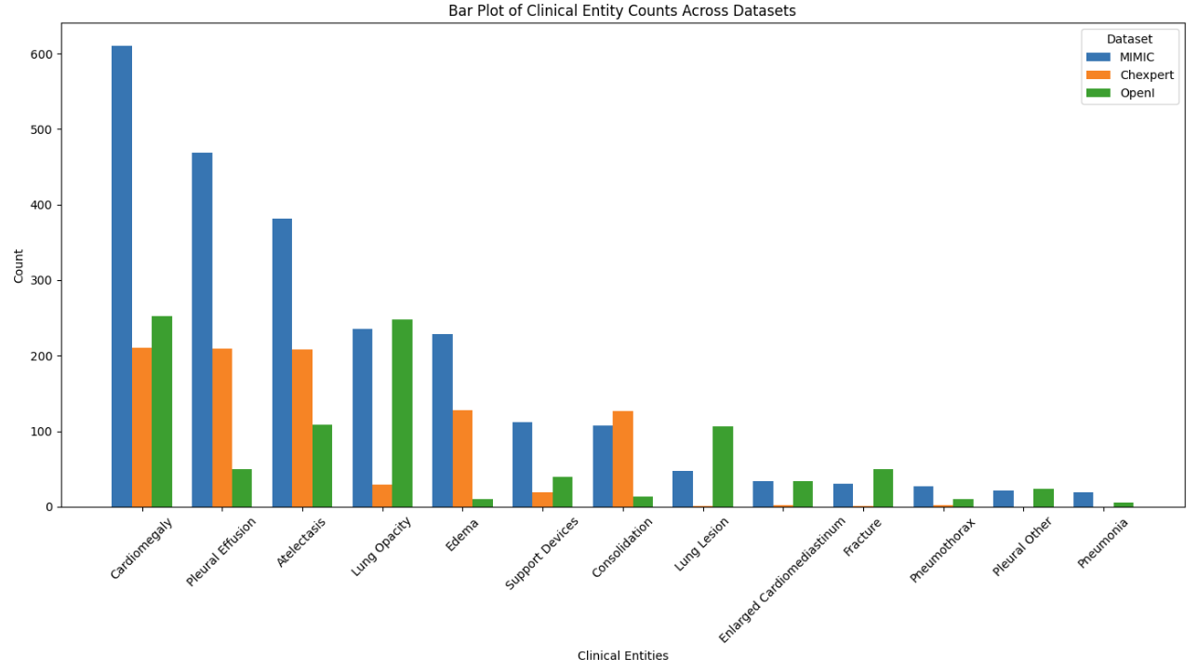}
    \caption{The number of selected entities in each dataset for CXR-Align.}
    \label{fig:bar}
\end{figure*}

The number of test samples for each dataset in CXR-Align is shown in \cref{table:dataset_count} and the distribution of selected entities is illustrated in \cref{fig:bar}. Entities are randomly selected with weights following the original distribution across test sets. Note that we prioritize cardiomegaly, atelectasis, edema, pleural effusion, pneumothorax, and consolidation, since the generated negations occur more often compared to other entities.

\begin{table}[h!]
\centering
\begin{tabular}{|c|c|c|}
\hline
\textbf{Dataset} & MIMIC & OpenI \\
\hline
\textbf{Count} & 2323 & 953 \\
\hline
\end{tabular}
\caption{Count of datasets used in the \textit{CXR-Align}.}
\label{table:dataset_count}
\end{table}

\begin{figure}[t]
    \centering
    \includegraphics[width=1\linewidth]{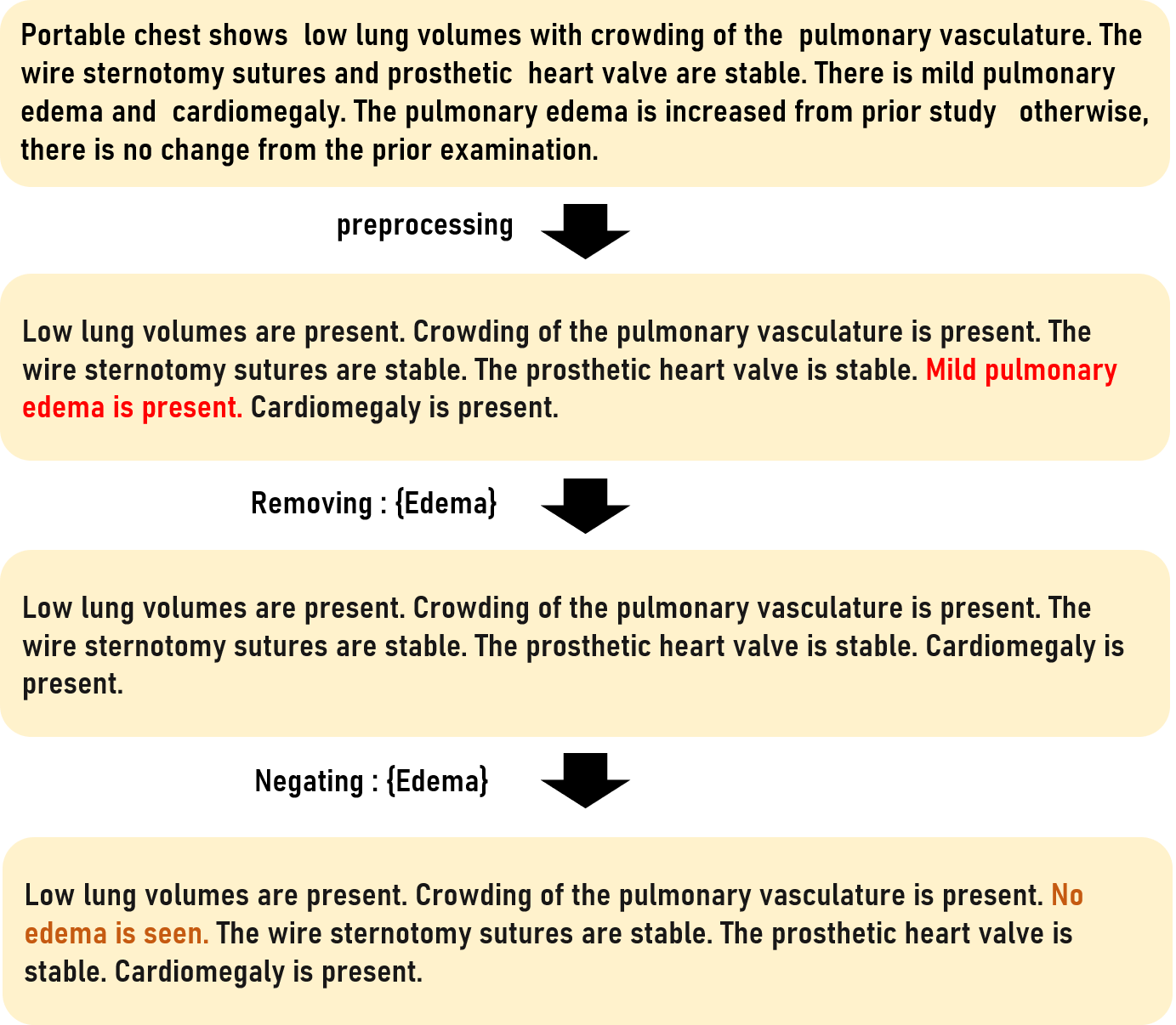}
    \caption{Example of the CXR-Align generation process.}
    \label{fig:exampleneg}
\end{figure}

\subsubsection{Process}
\label{sub:prompt_neg}
The process of CXR-Align generation is shown in \cref{fig:exampleneg}. The removal of findings is a very important step to avoid contradictions or inconsistencies within the report. When mediastinal-related finding is chosen, we add one of the following sentences into the report: 'The cardiomediastinal silhouette is normal.', 'The cardiac silhouette is unremarkable.', 'The heart size is normal.', 'The cardiomediastinal silhouette is within normal limits.', or 'No cardiomegaly.'. If other findings are chosen, we add one of the following templates: "No (finding) is seen.", "No (finding) is observed.", "There is no (finding).", or "No evidence of (finding).". Note that the negated sentence is inserted randomly within the report, either at the beginning, middle, or end. If all the sentences related to the finding were removed, we simply insert the negated statement.

\subsubsection{Prompts}
Below is the prompt for each step in LLM text preprocessing as in \cref{fig:cxrneg}.
\paragraph{Splitting}
We use the prompt from MAIRA2~\cite{maira2} for splitting reports so that each sentence represent and describe only one entity.
\paragraph{Removing Prior Reference}
"You are an expert chest X-ray (CXR) radiologist familiar with radiologic reports. Your task is to rewrite the given radiology reports by removing all references to prior reports or comparisons, while preserving the original structure as much as possible.
Input: A radiology report for a chest X-ray (CXR).
Output: A revised CXR report focusing solely on current medical findings, excluding references to prior reports, comparisons, and irrelevant details.
Guidelines:
Remove Comparisons: Eliminate any terms or phrases that suggest a comparison, such as "compared to," "in comparison with," "change", "cleared", "constant", "decrease", "elevate", "expand", "improve", "decrease", "increase", "persistent", "reduce", "remove", "resolve", "stable", "worse", "new", etc.
Focus on Current Findings: Ensure the report only describes the current state of the patient's lungs and related structures.
Preserve Medical Context: Maintain the original medical terminology and descriptions of abnormalities.
Retain Negations: Keep any negative statements about the absence of abnormalities.

Example 1:
Original: The left apex has not been included on this radiograph. The ET tube terminates 3.9 cm above the carina. The NG tube terminates in the stomach. Surgical clips and a faint metallic coil project over the chest. A left PICC terminates in the mid SVC. EKG leads overlie the chest wall. The lung volumes are low. There are persistent bilateral mid and lower zone hazy opacities. There are persistent bilateral hilar and perihilar linear opacities. No significant interval change is observed in the lung opacities. Bilateral pleural effusions are present. The right pleural effusion is greater than the left. No pneumothorax is observed on the right. No cardiomegaly is present. No interval change is observed in the mediastinal silhouette. No significant interval change is observed in the bony thorax.  
Revised: The left apex has not been included on this radiograph. The ET tube terminates 3.9 cm above the carina. The NG tube terminates in the stomach. Surgical clips and a faint metallic coil project over the chest. A left PICC terminates in the mid SVC. EKG leads overlie the chest wall. The lung volumes are low. There are persistent bilateral mid and lower zone hazy opacities. There are bilateral hilar and perihilar linear opacities. Bilateral pleural effusions are present. The right pleural effusion is greater than the left. No pneumothorax is observed on the right. No cardiomegaly is present. "

\paragraph{Omitting selected entity}
"Task: Given a specific finding or disease and a chest X-ray report, remove the sentences relevant to that finding or disease.

Context:

Lung lesion: Refers to nodule or mass.
Pleural other: Refers to pleural thickening.

Example:

Finding: Lung Lesion
Report: No pneumothorax is observed. No pleural effusion is observed. No evidence of hemorrhage is observed in the lung or mediastinum. Emphysema is severe. The heart size is normal. A complex of nodule and large bullae is present in the axillary region of the right upper lobe. 
Expected Output: No pneumothorax is observed. No pleural effusion is observed. No evidence of hemorrhage is observed in the lung or mediastinum. Emphysema is severe. The heart size is normal.

Finding: Cardiomegaly
Report: The feeding tube, with the wire stylet in place, is in the mid stomach. Heterogeneous pulmonary opacification is most pronounced in the left mid and lower lung. Heterogeneous pulmonary opacification is also present on the right, sparing only the upper lobe. The heart is mildly enlarged.
Expected Output: The feeding tube, with the wire stylet in place, is in the mid stomach. Heterogeneous pulmonary opacification is most pronounced in the left mid and lower lung. Heterogeneous pulmonary opacification is also present on the right, sparing only the upper lobe."

\subsection{Normal Case Detection} As described in \cref{ex:normalthinning}, we augmented the MIMIC dataset by adding 130,000 normal CXR images from a single tertiary hospital, each labeled with the report “No active lung lesion.” This augmentation results in an imbalanced dataset with 176,726 normal CXRs and 148,121 abnormal CXRs in the training set. For the reports which is used for the test set of this task, we included 2,999 abnormal reports sampled from the MIMIC test set with one normal report "No active lung lesion.". Data counts for the normal case detection experiment are provided in \cref{tab:data_summary_normalthinning}.
\begin{table}[h]
    \centering
    \begin{tabular}{l ccc}
        \toprule
        Dataset & Train & Valid & Test \\
        \midrule
        MIMIC-CXR   & 194,847 & - & - \\
        Private     & 130,000 & - & 1,026 \\
        \midrule
        Open-I      & -       & -     & 1,289 \\
        \bottomrule
    \end{tabular}
    \caption{Data counts for normal case detection experiment.}
    \label{tab:data_summary_normalthinning}
\end{table}

\subsection{Dataset Approvals and Ethics}

All procedures involving the MIMIC dataset, including large language model (LLM)-assisted report preprocessing and the construction of CXR-Align, were conducted in full compliance with PhysioNet’s guidelines for responsible LLM usage (\url{https://physionet.org/news/post/gpt-responsible-use}). Use of the private dataset was approved by the Institutional Review Board (IRB), with all researchers formally registered and authorized for data access. Due to licensing restrictions, only the MIMIC-based version of CXR-Align will be shared via PhysioNet, and access will be limited to credentialed users.

\section{Model}
\label{sec:model_info}
This section details the model implementation, augmentations, details with clinical information and hyperparameters.

\subsection{Implementation Details} The model is trained using the AdamW optimizer with a cosine learning rate schedule and linear warm-up. The learning rate is set to $4 \times 10^{-6}$, with a batch size of 64 over 10 epochs on a single A6000 GPU. For fine-tuning experiments, we set the learning rate to $1 \times 10^{-4}$, with a batch size of 128. We train for 200 epochs when fine-tuning with 10\% of the data, and for 20 epochs when fine-tuning with 100\% of the data, all on a single A6000 GPU. Each graph node's word is embedded using ClinicalBERT, and a one-hot code for class 'ANAT-DP', 'OBS-DP', 'OBS-DA', and 'OBS-U' is concatenated. The Graph Convolutional Neural Network (GCNN) for graph embeddings consists of two GCN conv layers with an input dimension of 772, a hidden dimension of 256, and an output dimension of 512 which is same with the other modalities.  The max token length is set to 300.

\subsection{Augmentations} For image augmentation, we apply Contrast Limited Adaptive Histogram Equalization (CLAHE) with a clip limit of 4, random resized cropping, and rotations of up to 10 degrees. Text augmentation consists of sentence shuffling only.

\subsection{Clinical Information} For the clinical information, we use CheXbert to extract the presence of entities. We additionally add one more label, where the value is 1 if all other labels are 0, and 0 otherwise. This accounts for cases where no findings are present, including entities that CheXbert may not cover. The entities are: [
        "Cardiomegaly", "Lung Opacity", "Atelectasis", "Lung Lesion", 
        "Pleural Effusion", "Fracture", "Support Devices", 
        "Enlarged Cardiomediastinum", "Pleural Other", 
        "Consolidation", "Edema", "Pneumothorax", "Pneumonia", "No Findings"].

\subsection{Hyperparameters}
\label{sub:hyper}
The temperature \(\tau\) is set to 0.1, and similarity thresholds for textual \(\tau_t\), clinical  \(\tau_c\), graph  \(\tau_g\) set at  0.9, 0.8, and 0.7, respectively. The weights for text \(w_T\), clinical \(w_C\), graph \(w_G\) weights in \cref{eq:final} are all set to 0.167.

\section{Evaluation Settings}
\subsection{zero-shot prompt}
Zero-shot prompt used for \cref{ex:main} is shown in \cref{table:findings}. For CheXpert multi-class classification, we follow the prompt used in CXR-CLIP. For adversarial prediction, we used the same prompts as in the "Others" category.
\begin{table}[h!]
\footnotesize
\centering
\begin{tabular}{|l|l|l|}
\hline
            & \textbf{Positive}                             & \textbf{Negative}                             \\ \hline
\textbf{RSNA}   & Findings suggesting pneumonia.               & No evidence of pneumonia.                     \\ \hline
\textbf{SIIM}   & There is pneumothorax                       & There is no pneumothorax                      \\ \hline
\textbf{Others} & There is \{findings\}                       & There is no \{findings\}                      \\ \hline
\end{tabular}
\caption{Positive and negative prompts for zero-shot evaluation.}
\label{table:findings}
\end{table}

\subsection{Report retrieval}
For report retrieval, we use the CheXbert F1 score rather than the standard BERTScore to evaluate how the retrieved or generated report clinically reflects the original report. The Macro F1 score is used since the Micro F1 score does not reflect the imbalance of the dataset.  Furthermore, rather than focusing on top-\(k\) retrieval performance, we emphasized clinical metrics because the test set contains reports with similar clinical semantics, which could bias the performance evaluation if based solely on top-\(k\) retrieval metrics.

\section{Additional Experiment}
\label{sec:sup_ex}

In this section, we provide a detailed discussion of our experiments. A notable finding from \cref{ex:main} is that our model's performance improves as we incorporate each similarity measure and hard negatives. Surprisingly, our baseline CLIP model's finetuned performance is comparable to or surpasses most of the SOTA CLIP models, implying that preprocessing steps like splitting reports and omitting prior references enhance the discriminability of CLIP models. Furthermore, adding similarity measures narrowed the gap between the RSNA and RSNA-\(ab\) results, indicating that our method helps the model to discriminate and correctly identify entities within abnormalities. In the following subsections, we provide a more detailed analysis of our benchmark \textit{CXR-Align}, adversarial prediction, normal case detection, and report retrieval.

\subsection{Detailed analysis on CXR-Align}
\begin{figure*}[t]
    \centering
    \includegraphics[width=0.8\textwidth]{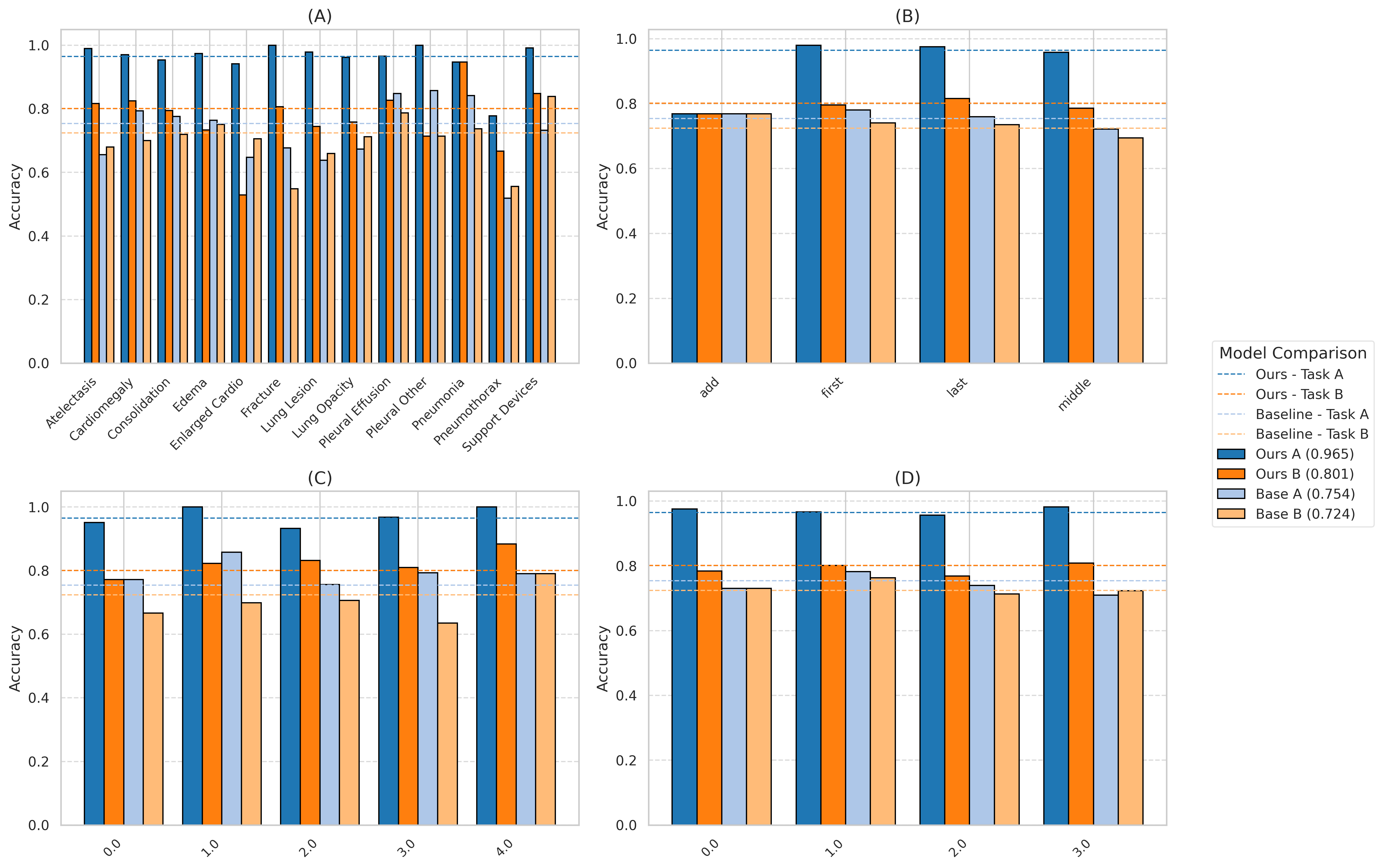}
    \caption{Detailed sub-analysis for CXR-Align on MIMIC dataset. (A) Task accuracy for entities that were either negated or removed. (B) Performance based on the location where the negated sentence was inserted. (C) Accuracy corresponding to the prompt used when the selected entity was related to mediastinal findings. (D) Performance corresponding to the prompt used when the selected entity was related to lung findings. For (C) and (D), refer to \cref{sub:prompt_neg}}
    \label{fig:mimicneg}
\end{figure*}

\begin{figure*}[t]
    \centering
    \includegraphics[width=0.8\textwidth]{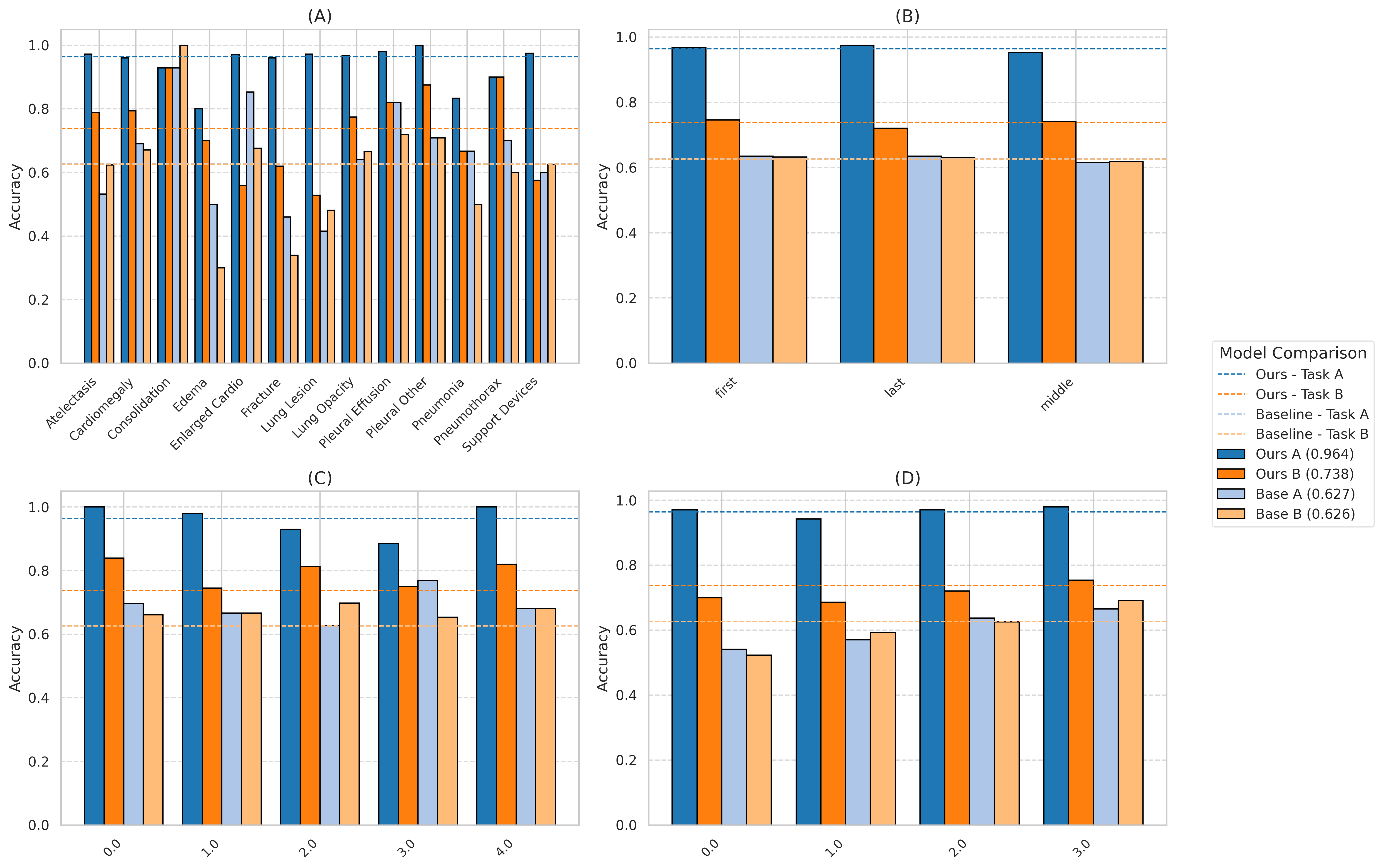}
    \caption{Detailed sub-analysis for CXR-Align on OPENI dataset. (A) Task accuracy for entities that were either negated or removed. (B) Performance based on the location where the negated sentence was inserted. (C) Accuracy corresponding to the prompt used when the selected entity was related to mediastinal findings. (D) Performance corresponding to the prompt used when the selected entity was related to lung findings. For (C) and (D), refer to \cref{sub:prompt_neg}}
    \label{fig:openineg}
\end{figure*}

\cref{fig:mimicneg} and \cref{fig:openineg} provide a detailed sub-analysis for the \textit{CXR-Align} benchmark on the MIMIC, and OpenI datasets, respectively. We analyze the following aspects:
\begin{enumerate}[leftmargin=*] \item \textbf{Entity Type}: For all datasets, negated entities related to 'pneumothorax', 'effusion', 'consolidation', 'enlarged cardiomediastinum', and 'pneumonia' performed below average, while the model best discriminated 'pleural other', 'support devices', and 'fracture'. This may be due to the prompts used to negate the latter entities being less frequent in the training set compared to the former. \item \textbf{Location}: The insertion location of the negation did not significantly affect performance, as accuracy was similar across all positions. \item \textbf{Mediastinal Prompt}: For prompts regarding mediastinal findings, Prompt 2 ('The heart size is normal') consistently resulted in below-average accuracy when inserted as a negated statement across all datasets. \item \textbf{Other Prompts}: For prompts related to lung entities, Prompt 2 ("There is no {finding}") performed the worst, falling below average. However, all prompts exhibited similar accuracy overall. \end{enumerate}

We hypothesize that the frequency of negated terms for each entity or prompt affects the model's performance and its comprehension of negations.

\subsection{Detailed analysis on Adversarial Prediction}
In this section, we perform a detailed analysis of adversarial prediction. We investigate how different models behave when subjected to this task compared to our model. As described in \cref{ex:adverserial}, this complex zero-shot task requires the model to determine whether one entity is present and another is absent. We conducted a total of 1,915 adversarial classification tasks. As shown in \cref{tab:adv_comparison} most SOTA models tend to predict an entity as positive when given an abnormal CXR, indicating that they do not effectively discern which entities are present or absent. This raises concerns about the zero-shot classification task discussed in \cref{ex:main} suggesting that models may focus on the overall abnormality of the CXR rather than understanding the full context and associating positivity with specific entities. While CXR-CLIP mitigated this issue to some extent, our model demonstrated better clinical understanding regarding the presence and absence of clinical findings.

\begin{table}[h]
    \centering
    \begin{tabular}{l c c c c}
        \toprule
        GT & \multicolumn{2}{c}{Present} & \multicolumn{2}{c}{Absent} \\
        \cmidrule(lr){2-3} \cmidrule(lr){4-5}
        Model & Positive & Negative & Positive & Negative \\
        \midrule
        GLORIA & 1671 & 244 & 1696 & 219 \\
        BioViL & 1539 & 376 & 1281 & 634 \\
        BioViL-T & 1625 & 290 & 1455 & 460 \\
        CXR-CLIP & 754 & 1161 & 341 & 1574 \\
        \midrule
        OURS & 720 & 1195 & 195 & 1720 \\
        \bottomrule
    \end{tabular}
    \caption{Positive/negative prediction counts in the adversarial prediction task for each model.}
    \label{tab:adv_comparison}
\end{table}

\subsection{Detailed Analysis on Normal Case Detection}
We conducted a detailed analysis of normal case detection, where the model must retrieve one normal report from 2,999 abnormal reports. As shown in \cref{tab:normal_skew},  training with long-tailed data containing more than 50\% normal CXR reports enables the model to effectively retrieve the normal report among all other abnormal reports. For the model trained only on the MIMIC dataset, the rank of the normal report was 68th. When using our internal test set as in \cref{tab:data_summary_normalthinning}, the model successfully retrieved the normal report with 99.7\% accuracy. This suggests that further training with internal data containing normal CXRs can achieve higher performance for internal tasks, allowing hospitals to build their own specialized models.

\begin{table}[h!]
\centering
\renewcommand{\arraystretch}{1.2} 
\begin{tabular}{|p{0.8\columnwidth}|r|}
\hline
\multicolumn{2}{|c|}{\textbf{OURS\(_{mimic}\)}} \\
\hline
There is a right lower lobe airspace consolidation. The lungs are otherwise clear. The hilar and cardiomediastinal contours are normal. There is no pneumothorax. There is no pleural effusion. Pulmonary vascularity is normal.  & 12 \\
\hline
A small residual area of linear atelectasis is present in the retrocardiac area. No pneumothorax is observed. No pleural effusion is observed. The heart size is normal. The hilar contours are normal. The mediastinal contours are normal. The visualized osseous structures are unremarkable. & 12 \\
\hline
The heart is normal in size. The mediastinal and hilar contours appear within normal limits. There is an inferolateral consolidation in the right upper lobe, consistent with pneumonia. The lungs appear clear elsewhere. No pleural effusions are present. No pneumothorax is present. The osseous structures are unremarkable. & 11 \\
\hline
\multicolumn{2}{|c|}{\textbf{OURS\(_{mimic+private}\)}} \\
\hline
No active lung lesion. & 1105 \\
\hline
No focal consolidation is seen. No pleural effusion is seen. No pneumothorax is seen. No pulmonary edema is seen. Minimal bronchial wall thickening is present. The heart size is normal. Mediastinal contours are normal. No bony abnormality is detected. & 48 \\
\hline
No lung consolidation. The left lower lung atelectatic band has resolved. Mediastinal and cardiac contours are normal. No pneumothorax. No pleural effusion. & 14 \\
\hline
\end{tabular}
\caption{Most frequent reports and their counts retrieved from the normal case detection task for the OpenI test images. The upper table shows results for our model trained only on MIMIC, while the lower table shows results for our model trained on MIMIC and private data.}
\label{tab:normal_skew}
\end{table}

\subsection{Report Retrieval}
\begin{figure*}[t]
    \centering
    \includegraphics[width=1\linewidth]{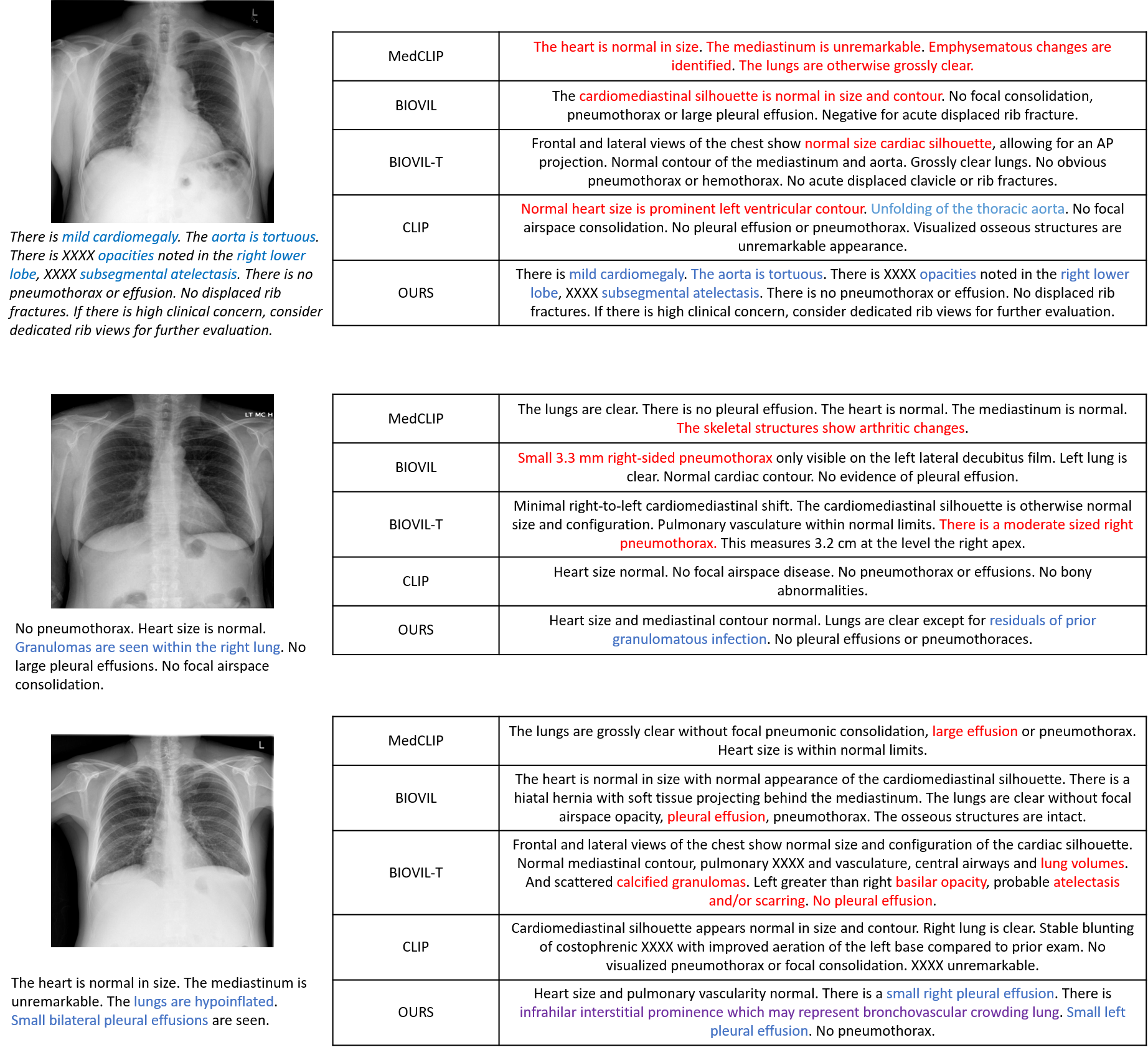}
    \caption{Examples of retrieved reports. Blue text represents important entities that should be included in the report. Red text indicates hallucinations or falsely interpreted entities. Purple represents clinically similar entities.}
    \label{fig:retrieved}
\end{figure*}

We provide examples of report retrieval performance in \cref{fig:retrieved}. Compared to other SOTA models and the baseline model, our model successfully retrieves reports that share similar semantics with the original report, even if they are not identical. Notably, in the third example, our model linked the textual semantics of "There is infrahilar interstitial prominence which may represent bronchovascular crowding lung" to the original report's "The lungs are hypoinflated," demonstrating high correlation.

\section*{Acknowledgments}  We would like to thank H.Y. Cho, I.H. Baek and Y.G. Kim for their valuable advice on this paper. This study was supported by the National Research Foundation of Korea (NRF) grants funded by the Ministry of Science and ICT (MSIT) (Grant No. RS-2024-00354666).

\section*{Comment on Manuscript Version}
This is the author-accepted manuscript of our paper accepted to the IEEE/CVF Conference on Computer Vision and Pattern Recognition (CVPR) 2025. This version includes an expanded ethics and data user agreement section, which was not part of the original submission to CVPR. The added section provides further transparency and compliance with data usage policies.

